\documentclass{article}

\usepackage{microtype}
\usepackage{graphicx}
\usepackage{subcaption}
\usepackage{ragged2e,array,booktabs}
\usepackage{enumitem}
\usepackage{multirow}
\usepackage{color, colortbl}
\usepackage{xspace}

\usepackage{hyperref}

\definecolor{Gray}{gray}{0.9}

\usepackage[accepted]{icml2024}

\usepackage{amsmath}
\usepackage{amssymb}
\usepackage{mathtools}
\usepackage{amsthm}
\usepackage{bm}

\usepackage[capitalize,noabbrev]{cleveref}

\theoremstyle{plain}

\theoremstyle{definition}

\theoremstyle{remark}

\newcommand{\sname}{BlobGEN\xspace}

\usepackage[textsize=tiny]{todonotes}

\icmltitlerunning{Compositional Text-to-Image Generation with Dense Blob Representations}

\begin{document}

\twocolumn[
\icmltitle{Compositional Text-to-Image Generation with Dense Blob Representations}




\begin{icmlauthorlist}
\icmlauthor{Weili Nie}{yyy}
\icmlauthor{Sifei Liu}{yyy}
\icmlauthor{Morteza Mardani}{yyy}
\icmlauthor{Chao Liu}{yyy}
\icmlauthor{Benjamin Eckart}{yyy}
\icmlauthor{Arash Vahdat}{yyy}
\end{icmlauthorlist}

\icmlaffiliation{yyy}{NVIDIA Corporation}

\icmlcorrespondingauthor{Weili Nie}{wnie@nvidia.com}


\vskip 0.3in
]



\printAffiliationsAndNotice{}  

\begin{abstract}

Existing text-to-image models struggle to follow complex text prompts, raising the need for extra grounding inputs for better controllability. In this work, we propose to decompose a scene into visual primitives -- denoted as dense blob representations -- that contain fine-grained details of the scene while being modular, human-interpretable, and easy-to-construct. Based on blob representations, we develop a blob-grounded text-to-image diffusion model, termed \sname, for compositional generation. Particularly, we introduce a new masked cross-attention module to disentangle the fusion between blob representations and visual features. To leverage the compositionality of large language models (LLMs), we introduce a new in-context learning approach to generate blob representations from text prompts. 
Our extensive experiments show that \sname achieves superior zero-shot generation quality and better layout-guided controllability on MS-COCO. When augmented by LLMs, our method exhibits superior numerical and spatial correctness on compositional image generation benchmarks. 
Project page: \url{https://blobgen-2d.github.io}.
\vspace{-0.4cm}
\end{abstract}

\begin{figure}[t]
    \centering
    \includegraphics[width=0.95\linewidth]{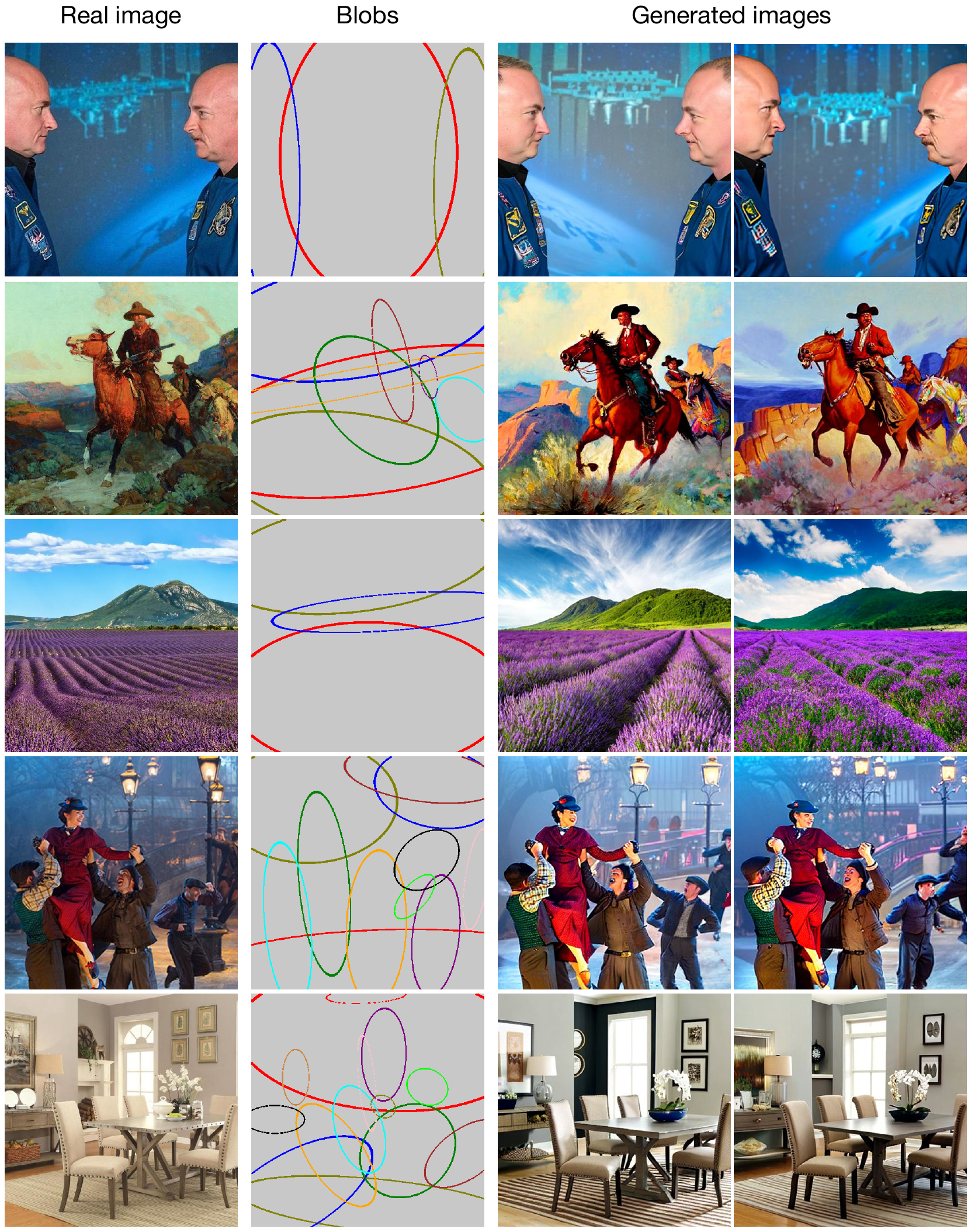}
    \vskip -0.1in
    \caption{
    Generated images from blob representations can reconstruct fine-grained details of real images. Each row shows the real image (Left), blobs (Middle), and two randomly generated samples (Right). We do not show blob descriptions for simplicity. }
    \label{fig:ll_const}
    \vspace{-0.4cm}
\end{figure}

\begin{figure*}[t]
    \centering
    \vspace{-0.2cm}
    \begin{subfigure}{0.34\textwidth}
        \centering
        \includegraphics[width=\linewidth,clip=True, trim={4.cm, 4.5cm, 11.0cm, 6cm}]{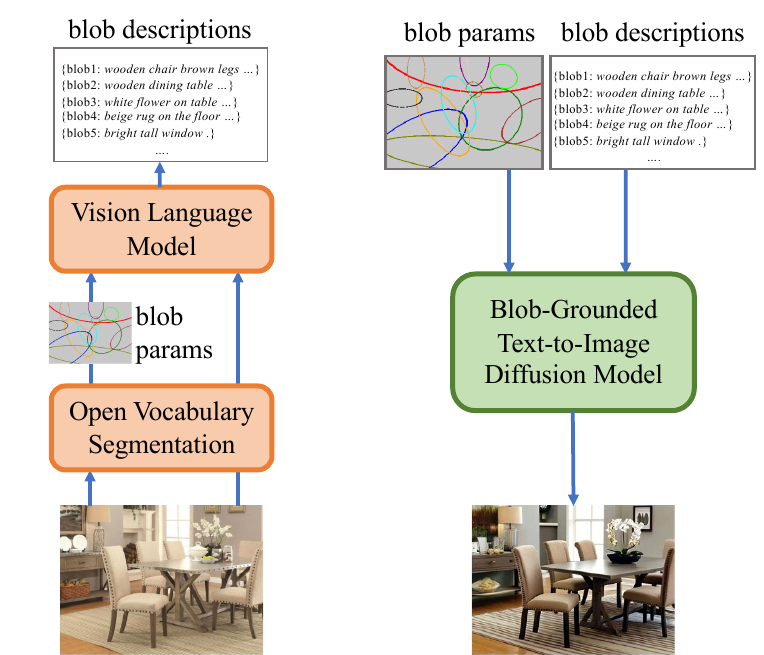}
        \caption{Blob representations}
        \label{fig:blob}
    \end{subfigure}%
    \hfill
    \begin{subfigure}{0.64\textwidth}
        \centering
        \includegraphics[width=\linewidth,clip=True, trim={0.cm, 4cm, 1cm, 3.5cm}]{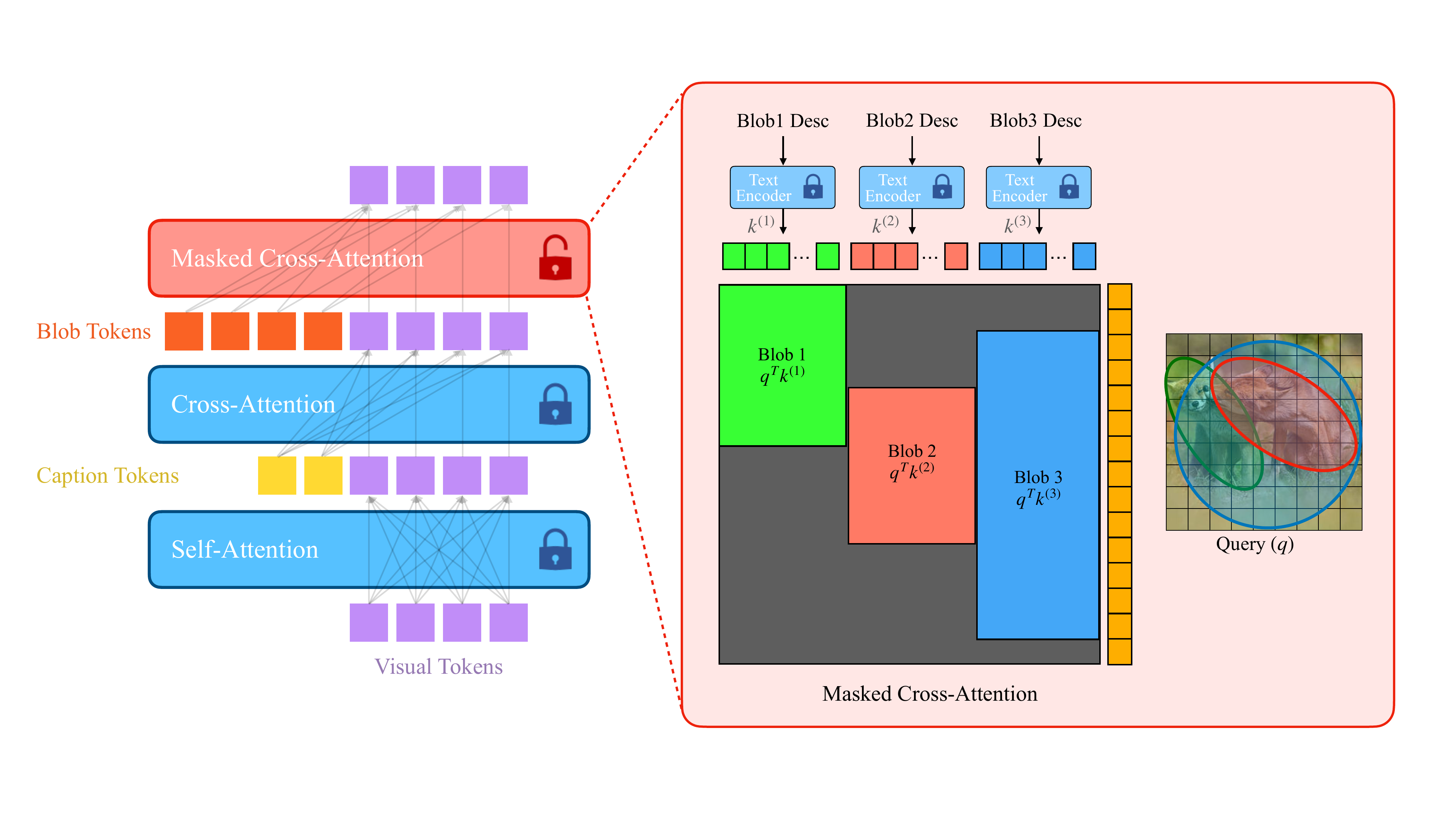}
        \caption{Masked cross-attention}
        \label{fig:masked_ca}
    \end{subfigure}%
    \vskip -0.15in
    \caption{(a) We extract blob representations (parameters and descriptions) using existing tools to guide the text-to-image diffusion model. (b) Our model leverages a novel masked cross-attention module that allows visual features to attend to only corresponding blobs.} 
    \label{fig:our_method}
    \vspace{-0.1cm}
\end{figure*}

\section{Introduction}
\label{intro}
Recent advances in text-to-image models enable us to generate realistic high-quality images~\citep{ramesh2022hierarchical,saharia2022photorealistic,podell2023sdxl,balaji2022ediffi}. This rapid rise in quality has been driven by new training and sampling strategies~\citep{ho2020denoising,song2020score}, new network architectures~\citep{dhariwal2021diffusion,rombach2022high}, and internet-scale image-text paired data~\citep{schuhmann2022laion}.
Despite the progress, current large-scale text-to-image models struggle to follow complex prompts, where they tend to misunderstand the context and ignore keywords~\citep{betker2023improving,huang2023t2i}. 
Thus, fine-grained controllability is an open problem.

To cope with these challenges, recent works have attempted to condition text-to-image models on visual layouts. 
Since a text prompt can be vague in describing visual concepts (\emph{i.e.}, the precise location of an object), image generation models may face difficulty striking a balance between expressing the given information and hallucinating missing information. Additional grounding inputs can guide the generation process for better controllability. 
These layouts can be represented by bounding boxes~\citep{li2023gligen}, semantic maps~\citep{zhang2023adding}, depths~\citep{rombach2022high}, and other modalities~\citep{li2023gligen,zhang2023adding}. 
Among them, semantic and depth maps provide fine-grained information but are not easy for users to construct and manipulate. In contrast, bounding boxes are user-friendly but contain more coarse-grained information than semantic and depth maps~\citep{wang2016deep,li2023gligen}. 


In this work, we introduce a new type of visual layout, termed \textit{dense blob representation}, to serve as grounding inputs to guide text-to-image generation. The blob representations correspond to visual primitives (\emph{i.e.}, objects in a scene) and can be automatically extracted from a scene. Specifically, a blob representation has two components: 1) the blob parameter, which formulates a tilted ellipse to specify the object's position, size and orientation; and 2) the blob description, which is a rich text sentence that describes the object's appearance, style, and visual attributes. With this definition, our proposed blob representation can largely preserve the fine-grained layout and semantic information of a scene (see Figure~\ref{fig:ll_const}). Furthermore, since blob parameters and descriptions are both represented with structured texts, they can be easily constructed and manipulated by users. 

We then develop a blob-grounded text-to-image diffusion model, termed \sname, that is built upon existing diffusion models, with blob representations as grounding input.  
To disentangle the fusion between blob representations and visual features,
we devise a masked cross-attention module that relates each blob to the corresponding visual feature solely in its local region.
Furthermore, inspired by \citet{feng2023layoutgpt,lian2023llmgrounded} who prompt LLMs to plan box layouts, 
we design a new in-context learning approach for LLMs to generate dense blob representations from text prompts. By augmenting our model with LLMs, we can leverage 
the visual understanding and compositional reasoning capabilities
of LLMs to solve complex compositional image generation tasks.
Our model paves the way for a modular framework where images can be easily generated or manipulated by users and LLMs.

Our extensive experiments indicate that \sname achieves superior zero-shot generation quality on MS-COCO~\citep{lin2014microsoft}. For instance, it improves the zero-shot FID of base model from 10.40 to 8.61, and offers much better layout-guided controllability than GLIGEN~\citep{li2023gligen} as demonstrated by region-level CLIP~\citep{radford2021learning} scores. By solely modifying a single blob representation while holding other blobs static, \sname exhibits a strong local editing and object repositioning capability. With LLM augmentation, our method excels in compositional generation tasks. For instance, our method outperforms LayoutGPT~\citep{feng2023layoutgpt} by 5.7\% and 1.4\% for spatial and numerical accuracy on NSR-1K~\citep{feng2023layoutgpt}.

Overall, our main contributions are summarized as follows:
\vspace{-8pt}
\begin{itemize}[leftmargin=*]\setlength\itemsep{-0em}
    \item We propose to decompose a scene into dense blob representations, each of which represents fine-grained details of a visual primitive (\emph{i.e.}, an object) in the scene.
    \item We propose \sname, a blob-grounded modular text-to-image model with a new masked cross-attention module that takes blob representations as grounding input.
    \item We augment our model with LLMs for compositional generation, by designing a new in-context learning approach for LLMs to infer blob representations from text prompts.
    \item We show our method achieves better zero-shot generation performance on MS-COCO, and has better numerical and spatial correctness in compositional benchmarks.
\end{itemize}




\section{Method}
\label{method}
We first introduce our image decomposition into blob representations, and then describe the new generative framework that conditions on blob representations to generate images. Finally, we present the customized in-context learning procedure that prompts LLMs to generate blobs.

\subsection{Image Decomposition into Blob Representations}
Given an image, we aim to extract visual primitives or object-level representations that satisfy two properties: 1) They contain fine-grained details of the scene such that the original image  can be \textit{semantically} reconstructed in the maximum degree from them, and 2) they are modular, human-interpretable and easy to construct or manipulate, which means users can create and edit an image efficiently.
To this end, we propose a new type of visual layouts, termed \emph{dense blob representations}, each of which describes a single object in a scene. A blob representation consists of two components: \emph{blob parameter} and \emph{blob description}. 

Formally, a blob parameter specifies the size, location, and orientation of the blob using a vector of five variables $[c_x, c_y, a, b, \theta]$, where ($c_x$, $c_y$) is the center point of the ellipse, $a$ and $b$ are the radii of its semi-major and semi-minor axes, and $\theta \in (-\pi, \pi]$ is the orientation angle of the ellipse.
Intuitively, similar to the functionality of bounding boxes~\citep{li2023gligen,yang2023reco}, the blob parameter can represent the location and size of an object. On the other hand, due to the existence of the orientation angle $\theta$, the visual layout depicted by a blob parameter is more fine-grained than a bounding box: 1) it can additionally describe the orientation or pose of an object, and 2) it can more precisely describe the shape and size of an object, particularly those with an elongated shape and a large inclined angle.

A blob description is a text sentence that describes the visual appearance of an object, complementing the spatial layout information depicted by the blob parameter. In this work, we use a region-level synthetic caption extracted by a pre-trained image captioner as the blob description. As shown in Figure~\ref{fig:blob}, it not only provides the category name but also captures the detailed visual features of an object, including its appearance (\emph{e.g.}, color, texture, and material, etc.) and the spatial relationship of sub-parts within the object region (\emph{e.g.}, ``a wooden chair with brown legs and soft seat''). 

Since our blob representations retain the fine-grained visual layouts and other detailed visual features of the original image, it can faithfully recover the image with a diffusion model (see Figure~\ref{fig:ll_const}). Note that blob representations can also capture irregular, large objects and background (see Figure~\ref{fig:blobgen_irregular_large}).
Moreover, both blob parameters and descriptions are in the form of simple text inputs, and thus they can be easily constructed and manipulated by human users and even generated by LLMs as we will show next. 


\subsection{Blob-grounded Text-to-Image Generation}

Existing text-to-image diffusion models often consist of convolutional and self-attention layers that operate on image features directly, and cross-attention layers that inject text conditioning into the network (see Figure~\ref{fig:masked_ca}).
We build \sname upon the pre-trained text-to-image Stable Diffusion model, where we introduce new cross-attention layers to incorporate blob grounding into the diffusion model. To retain the prior knowledge of pre-trained models for synthesizing high-quality images, we freeze their weights and only train the newly added layers. In the following, we highlight 
the key design choices for blob-grounded generation.

\paragraph{Blob Embedding. }
Denote the blob parameter as $\pmb{\tau}: = [c_x, c_y, a, b, \theta]$ and blob description as $\pmb{s} := [\pmb{s}_1, \cdots, \pmb{s}_L]$, where $L$ is the text sentence length. For blob parameter $\pmb{\tau}$, we first encode its orientation angle $\theta$ to the sine and cosine representation $(\sin{\theta}, \cos{\theta})$, and then obtain the blob parameter embedding $\pmb{e}_\tau = \text{Fourier}(\tilde{\pmb{\tau}}) \in \mathbb{R}^{d_\tau}$, where $\tilde{\pmb{\tau}} : = [c_x, c_y, a, b, \sin{\theta}, \cos{\theta}]$ and $\text{Fourier}(\cdot)$ denotes the Fourier feature encoding~\citep{tancik2020fourier}. For blob description, we use the CLIP text encoder $f$ to obtain the sentence embedding $\pmb{e}_s = f(\pmb{s}) := [\pmb{e}_{s_1}, \cdots, \pmb{e}_{s_L}] \in \mathbb{R}^{L \times d_s}$. 
Before we pass the blob sentence embedding to the network, we first concatenate the two embeddings $\pmb{e}_\tau$ and $\pmb{e}_s$. Thus, the final blob embedding is given by
$$\pmb{e}_b = \text{MLP}([\tilde{\pmb{e}}_{s_1}, \cdots, \tilde{\pmb{e}}_{s_L}]) \in \mathbb{R}^{L \times d_b}$$
where $\tilde{\pmb{e}}_{s_l} := [\pmb{e}_{s_l}; \pmb{e}_{\tau}] \in \mathbb{R}^{d_s + d_\tau}$ for all $l \in \{1,\cdots, L\}$ with $[\cdot;\cdot]$ denoting a concatenation along the feature dimension, and $\text{MLP}(\cdot)$ represents an MLP layer. 

\paragraph{Masked Cross-Attention.}
Given $N$ blob embeddings, denoted as $\{\pmb{e}_b^{(n)}\}_{n=1}^N$, we represent $\pmb{g} \in \mathbb{R}^{hw\times d_g}$ as the visual features of an image, where $h$ and $w$ represent the spatial size of the feature maps. If the query, key and value are denoted by $\pmb{q} := \pmb{g}\pmb{W}_q \in \mathbb{R}^{hw \times d_g}$, $\pmb{k}^{(n)} := \pmb{e}_b^{(n)}\pmb{W}_k^{(n)} \in \mathbb{R}^{L \times d_g}$, and $\pmb{v}^{(n)} := \pmb{e}_b^{(n)}\pmb{W}_v^{(n)} \in \mathbb{R}^{L \times d_g}$, respectively, a standard cross-attention between $\pmb{g}$ and $\{\pmb{e}_b^{(n)}\}_{n=1}^N$ is 
\begin{align*}
\small
    \text{CA}(\pmb{g}, \{\pmb{e}_b^{(i)}\}) = \sigma(\frac{\pmb{q}[\pmb{k}^{(1)}; \cdots; \pmb{k}^{(N)}]^T}{\sqrt{d_g}}) [\pmb{v}^{(1)}; \cdots; \pmb{v}^{(N)}]
\end{align*}
where $[\cdot; \cdot]$ is a concatenation along the sequence dimension and $\sigma(\cdot)$ is the softmax function. 
We can see that, in the standard cross-attention, every blob embedding attends to every feature ``pixel'' (in the $h \times w$ plane) of the feature maps. This is undesirable since blob embedding only conveys information about its corresponding local region, and its interaction with other regions may confuse the model, leading to more text leakage and entanglement in generation. 

To solve this issue, we propose to mask the feature maps $\pmb{g}$ such that each blob embedding only attends to its local region, as shown in Figure~\ref{fig:masked_ca}. Denote the attention mask for the $i$-th blob as $\pmb{m}^{(i)} \in \mathbb{R}^{hw}$. It is obtained by downsampling the $i$-th blob's binary ellipse mask 
where a pixel value is 1 if it is within the blob ellipse, and 0 otherwise.
Accordingly, we define the \emph{masked cross-attention} as
\begin{align*}
\small
    \text{CA}_{m}(\pmb{g}, \{ \pmb{e}_b^{(i)}, \pmb{m}^{(i)} \}) = \sigma(\frac{[\pmb{a}^{(1)}; \cdots; \pmb{a}^{(N)}]}{\sqrt{d_g}}) [\pmb{v}^{(1)}; \cdots; \pmb{v}^{(N)}]
\end{align*}
where the $i^{th}$ attention weight for the $j^{th}$ location is:
%
\begin{align*}
    \pmb{a}^{(i)}_j = 
    \begin{cases}
    \pmb{q}_j \pmb{k}^{(i)T} \ &\text{if } \pmb{m}^{(i)}_j = 1  \\
     -\infty &\text{otherwise}
  \end{cases} \ \ \text{for } j \in \{1, 2, \dots, hw\}.
\end{align*}
With this masking design, blob representations and local visual features are well aligned in an explicit manner. Therefore, the blob grounding process can be more modular and independent across different object regions, and the model can be more disentangled in generation. 

\paragraph{Other Design Choices. }
Similar to \citep{alayrac2022flamingo,li2023gligen}, we also add the masked cross-attention module in a gated way, where a learnable scalar controls the information flow from the cross-attention branch for more stable training. We optionally add the gated self-attention module proposed by \cite{li2023gligen}, which shows a slight improvement in generation quality. We do not make any changes to self-attention and convolutional layers, allowing information to propagate across different image regions for overall long-range correlations. For the image-level global captions, we find synthetic ones work better than original real captions~\cite{betker2023improving}, so we only use synthetic global captions to train our model (see Appendix~\ref{sec:syn_caption_appendix}). Finally, we use the original denoising score matching loss~\citep{ho2020denoising} to train only the new parameters.

\vspace{-3pt}
\subsection{LLMs for Blob Generation}
\vspace{-1pt}

Here, we aim to show that our blob representations can be generated by LLMs. 
Specifically, we design two separate in-context learning processes: one for generating blob parameters and another for generating blob descriptions.

\vspace{-3pt}
\paragraph{Blob Parameter Generation.}
Inspired by \citet{feng2023layoutgpt}, we adopt the CSS format to represent blob parameters such that LLMs better understand their spatial meaning. Each generated layout in an in-context example starts with the category name, followed by a declaration section in the CSS style, which is \texttt{"object \{major-radius: ?px; minor-radius: ?px; cx: ?px; cy: ?px; angle: ?\}"}. The first four values are measured in pixel length, whereas the last value for angle is expressed in degree and normalized to be within $[0, 180]$. All values are rounded to integers. Next, we follow the procedure of \citet{feng2023layoutgpt} to select top-$k$ similar demonstration examples\footnote{In fact, retrieval from a large blob dataset to obtain in-context demonstration examples is \textit{not necessary} for our method. See Appendix~\ref{sec:wo_retrieval} for more details.}. The final prompt for LLMs consists of a system prompt that instructs the blob parameter generation, $k$ demonstration examples, and the test prompt (usually a global caption).  See Appendix~\ref{sec:blob_param_appendix} for details. 

\vspace{-3pt}
\paragraph{Blob Description Generation.}
Blob descriptions are less structured as they are essentially a list of text sentences. Thus, we do not use the CSS format to generate them but we still use the category name as a separator between blobs for the ease of LLM generation. Thus, each generated blob description in an in-context example is formatted as \texttt{"object \{text sentence\}"}. We utilize the same method to select top-$k$ demonstration examples and construct the final prompt, which includes a system prompt that instructs the blob description generation, $k$ demonstration examples, and the test prompt. See Appendix~\ref{sec:blob_desc_appendix} for details. 




\begin{table*}[t]
\vspace{-0.2cm}
\caption{Evaluation of zero-shot generation quality and layout-guided controllability on MS-COCO validation set. $^\dagger$These models need to use an extra database for retrieval-augmented generation. }
\label{tab:gen_contr}
\vskip -0.2in
\begin{center}
\begin{small}
\resizebox{0.75\textwidth}{!}{
\begin{tabular}{lcccccc}
\toprule
\multirow{2}{*}{Method} & \multirow{2}{*}{\#Parameters} & Zero-shot Generation & \multicolumn{3}{c}{Controllability} \\ \cmidrule(lr){3-3} \cmidrule(lr){4-6} & & FID~$\downarrow$ & mIOU~$\uparrow$ & rCLIP$_t$~$\uparrow$  & rCLIP$_i$~$\uparrow$ \\
\midrule
CogView~\citep{ding2021cogview}   & 4B & 27.10 & - & - & - \\
KNN-Diffusion~\citep{sheynin2022knn} & 470M$^\dagger$ & 16.66 & - & - & - \\
GLIDE~\citep{sheynin2022knn} & 6B & 12.24 & - & - & - \\
Make-a-Scene~\citep{gafni2022make} & 4B  & 11.84 & - & - & - \\
DALL-E 2~\citep{ramesh2022hierarchical}  & 5.5B  & 10.39 & - & - & - \\
\textsc{Lafite}2~\citep{zhou2022lafite2} & 1.45B$^\dagger$  & 8.42 & - & - & - \\
Muse~\citep{chang2023muse}  & 3B   & 7.88 & - & - & - \\
Imagen~\citep{saharia2022photorealistic}  & 8B  & 7.27 & - & - & -         \\
Parti~\citep{yu2022scaling}   & 20B   & 7.23 & - & - & - \\
Re-Imagen~\citep{chen2022re}  & 8B$^\dagger$   & 6.88 & - & - & - \\
\midrule
\rowcolor{Gray}
\emph{w/ SD decoder~\citep{rombach2022high}} & & & & & \\ 
Base model (SD-1.4, ~\citealt{rombach2022high})  & 1B    & 11.14 & 0.1338 &	0.2443	& 0.7558       \\
GLIGEN~\citep{li2023gligen}  & 1.2B    & 11.63 &	 0.4154	& 0.2688	& 0.7941   \\
GLIGEN w/ synthetic captions  & 1.2B    & 10.80	 & 0.4143 &	0.2724	& 0.7964       \\
Ours  & 1.4B    & 8.94	& \underline{0.5000}	& \textbf{0.2906}	& \underline{0.8241}       \\
\midrule
\rowcolor{Gray}
\emph{w/ consistency decoder~\citep{betker2023improving}} & & & & & \\ 
Base model (SD-1.4, ~\citealt{rombach2022high})  & 1.5B    & 10.40 &	0.1316	& 0.2381 &	0.7520      \\
GLIGEN~\citep{li2023gligen}  & 1.7B    & 11.15	& 0.4238	& 0.2591	& 0.7976      \\
GLIGEN w/ synthetic captions & 1.7B  & 10.54	& 0.4244	& 0.2609	& 0.7994       \\
Ours  & 1.9B    & 8.61 &	\textbf{0.5103}	& \underline{0.2794}	& \textbf{0.8288}       \\
\bottomrule
\end{tabular}}
\end{small}
\end{center}
\vskip -0.1in
\end{table*}

\vspace{-3pt}
\section{Related Work}
\label{rw}
\vspace{-1pt}

\paragraph{Text-to-Image Generation.}
Large text-to-image generative models have attracted much attention in the past few years, due to their unprecedented photorealism in generation. Among them, many methods are based on diffusion models~\citep{ramesh2022hierarchical,rombach2022high,saharia2022photorealistic,balaji2022ediffi} or auto-regressive models~\citep{ramesh2021zero,yu2022scaling,chang2023muse}.
Instead of purely improving visual quality, recent models aim at improving their prompt following capabilities~\citep{podell2023sdxl,betker2023improving}, where training on synthetic image captions becomes a promising direction~\citep{betker2023improving,chen2023pixart,wu2023paragraph}. Different from them, we use the synthetic caption as an object-level text description for each blob.
By conditioning existing text-to-image models on blob representations, our generation can follow more fine-grained, object-level user instructions while maintaining high visual quality.

\paragraph{Compositional Image Generation.}
Early works have proposed to learn concept distributions defined by energy functions, which can be explicitly combined~\citep{du2020compositional,nie2021controllable,liu2022compositional}. Blob parameters have also been used for spatially disentangled generation with GANs~\citep{epstein2022blobgan}. Based on text-to-image models, recent methods have focused on learning special text tokens to represent concepts and injecting them into the text prompt for concept compositions~\citep{ruiz2023dreambooth,kumari2022customdiffusion,xiao2023fastcomposer}. Other methods have explored guiding internal representations through cross-attention maps to steer sampling~\citep{feng2022training,chefer2023attend,epstein2023diffusion,chen2023trainingfree,phung2023grounded}. 
Another line of research aims at conditioning text-to-image models on extra visual layouts for better controllability~\citep{yang2023reco,li2023gligen,zheng2023layoutdiffusion,huang2023composer,feng2023ranni}. 
However, none of them uses blob representations as grounding inputs for compositional generation. 

\paragraph{LLM-augmented Image Generation.}
With their generalization abilities, LLMs have also been used in text-to-image generation. \citet{wu2023visual} introduce a prompt manager that links LLMs with various text-to-image models to execute complex image synthesis tasks. Several works propose to fuse LLMs with text-to-image models for various multimodal generation tasks by mapping between their embedding spaces~\citep{koh2023generating,sun2023generative}. More similarly, other works use LLMs to infer visual layouts from text prompts as grounding inputs for text-to-image models~\citep{feng2023layoutgpt,lian2023llmgrounded,feng2023ranni}. They demonstrate that LLMs can generate bounding boxes and well-structured attributes with carefully designed prompts, but it remains unclear whether blob representations can be generated by LLMs and how robust our blob-grounded generative model is to LLM-planned layouts.

\vspace{-3pt}
\section{Experiments}
\label{exp}
\vspace{-1pt}

We first evaluate the generation performance of our blob-grounded text-to-image generative model; following that, we evaluate its compositional reasoning performance when augmented with in-context LLMs for blob generation. 

\subsection{Blob-grounded Text-to-Image Generation}

Here we compare BlobGEN with previous methods on zero-shot image generation, and perform ablation studies to highlight the impact of each design choice in our method.

\subsubsection{Experiment Setup}

\paragraph{Data Preparation.}
We use a dataset of random 1M image-text pairs from the Common Crawl web index (filtered with the CLIP score) and resize all images to a resolution of 512$\times$512. To extract blob representations for each image, we first apply ODISE~\citep{xu2023open} to get instance segmentation maps, followed by an ellipse fitting optimization to determine blob parameters for each map, aiming to maximize the Intersection Over Union (IOU) between the blob ellipse and segmentation mask. With segmentation maps, we crop out local regions for all objects in an image and use LLaVA-1.5~\citep{liu2023improved} to caption each blob. On average, each image contains 12 blobs. 

\paragraph{Training Details.}
Our model adopts the LDM framework~\citep{rombach2022high} and is built upon the SD-1.4 checkpoint. An image of resolution 512$\times$512 is mapped to a latent space of $64\times64\times4$ by an image encoder. By default, our model is trained on 1M samples for 400K steps using a batch size of 512, requiring 9 days on 64 NVIDIA A100 GPUs. We use the AdamW optimizer~\citep{loshchilov2018decoupled} and the learning rate of $5\times10^{-5}$ with a linear warm-up for the first 10K steps. We set the maximum number of blobs per image to 15. To encourage the model to rely more strongly on the blob representations, we randomly drop the global caption with $50\%$ probability.

\paragraph{Evaluation Metrics.}
We use FID~\citep{heusel2017gans} to compare the visual quality of generated images from different models. Unless stated otherwise, all FIDs are computed with 30K generated and real images. To the measure the controllability, i.e., how well the generation follows the layout guidance, we propose three metrics: mIOU, rCLIP$_i$ and rCLIP$_t$. The mIOU is defined as the mean IOU between the segmentation maps by applying LangSAM\footnote{https://github.com/luca-medeiros/lang-segment-anything.} to the generated image and the region ellipse masks depicted by input blob parameters. The rCLIP$_i$ is defined as the region-level CLIP score between the generated image and the ground-truth real image (paired with the input global caption). Here, ``region-level'' means that we pass a cropped image specified by the blob parameter to the CLIP image encoder for an image embedding. Similarly, the rCLIP$_t$ is defined as the region-level CLIP score between the generated image and the corresponding blob descriptions.

\begin{figure}[t]
    \centering
    \includegraphics[width=0.98\linewidth]{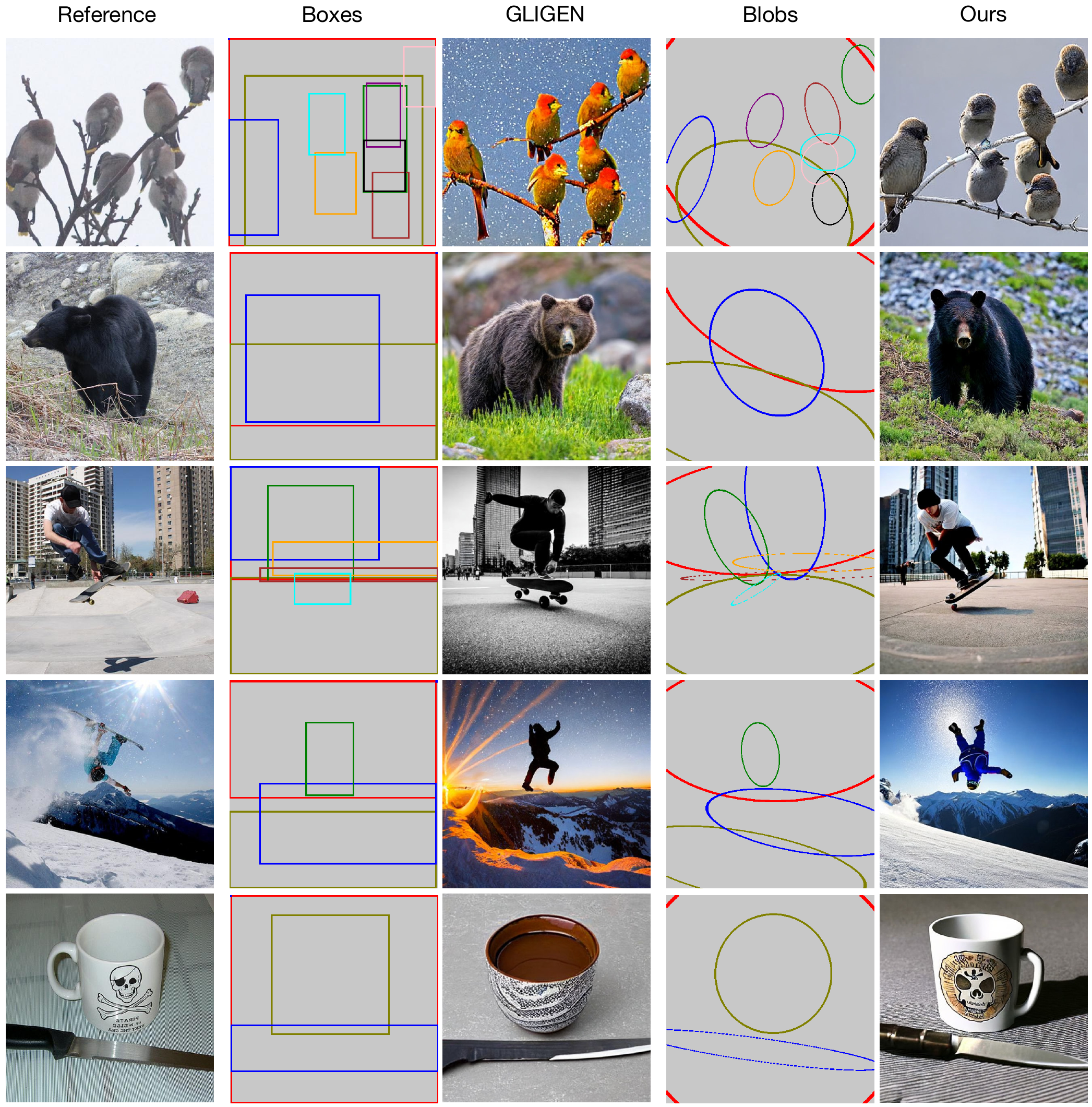}
    \vskip -0.1in
    \caption{Zero-shot layout-grounded generation results of GLIGEN and our method on MS-COCO validation set. In each row, we visualize the reference real image (Left), bounding boxes and GLIGEN generated image (Middle), blobs and our generated image (Right). All images are in resolution of 512$\times$512. }
    \label{fig:coco_gen}
    \vspace{-5pt}
\end{figure}

\begin{figure*}
    \centering
    \includegraphics[width=0.98\textwidth]{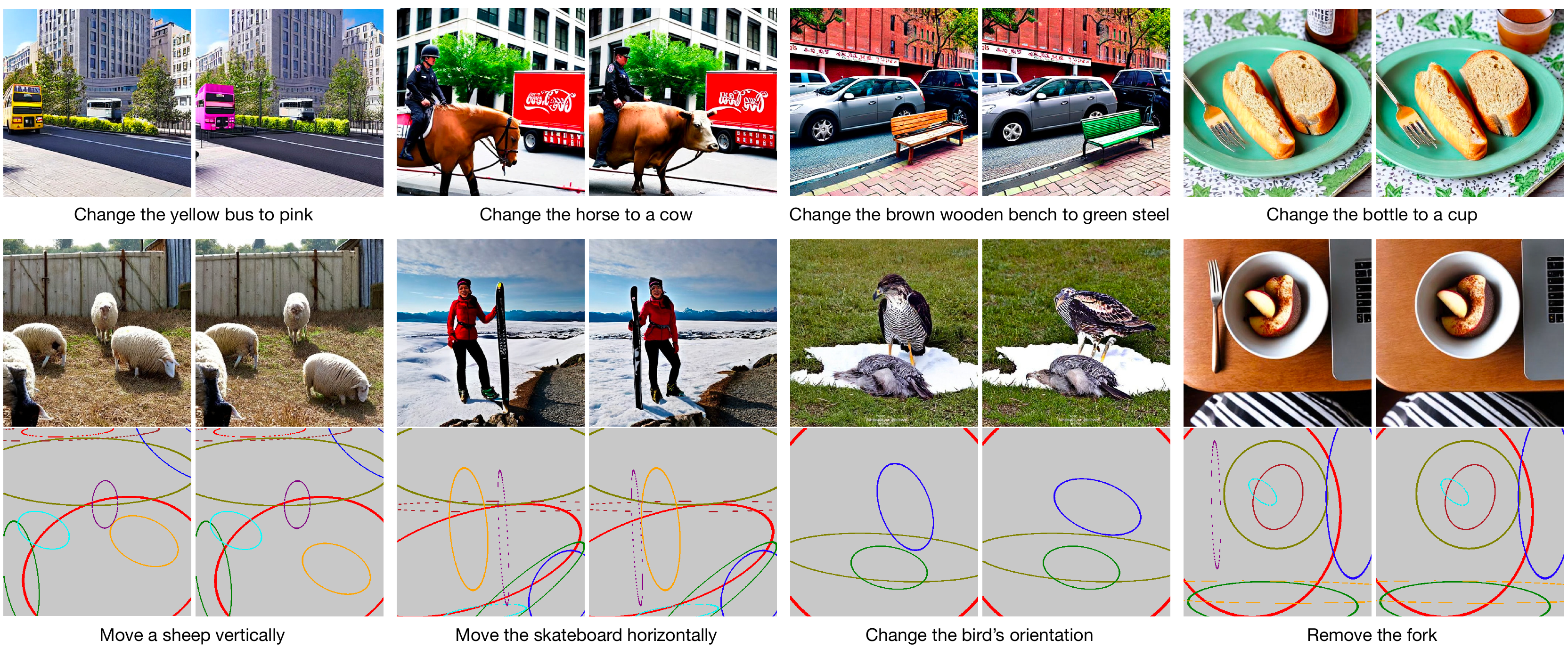}
    \vskip -0.1in
    \caption{Various image editing results of our method on the MS-COCO validation set, where each example contains two generated images: (Left) original setting and (Right) after editing. The top row shows the local editing results where we only change the blob description and since the blob parameters stay the same after editing, we do not show blob visualizations. The bottom two rows show the object reposition results where we only change the blob parameter. All images are in resolution of 512$\times$512. }
    \label{fig:coco_edit}
    \vspace{-1pt}
\end{figure*}

\subsubsection{Zero-shot Generation on MS-COCO }

\paragraph{Quantitative Results.}
In Table~\ref{tab:gen_contr}, we compare our method with the state-of-the-art models in terms of zero-shot generation quality and controllability on the MS-COCO validation set. For our method and two closely related baselines (base model and GLIGEN), we report the results with different image decoders (\emph{i.e.}, SD decoder and consistency decoder). First, we observe that our method achieves lower FID than both GLIGEN, which uses bounding boxes as grounding input, and the base model SD-1.4. The consistency decoder always improves FID but it also increases the overall model size. When compared with other text-to-image models that have a much larger model size, our FID also remains competitive. It implies that adding blob representations largely improves the image synthesis quality. 

Furthermore, our method outperforms both GLIGEN and base model by a large margin regarding all three metrics for controllability: mIOU, rCLIP$_i$ and rCLIP$_t$. This demonstrates that our generated images also have better region-level consistency with the grounding inputs, besides better image-level visual quality. Also, the consistency decoder in general achieves better mIOU and rCLIP$_i$ than the SD decoder, but it deteriorates the rCLIP$_t$ score. We hypothesize that this discrepancy arises from a misalignment between the consistency decoder and the CLIP text encoder. GLIGEN trained with synthetic captions can also slightly improve the controllability scores over its counterpart with original real captions. This further supports the recent finding that training on higher-quality synthetic captions can improve the model's prompt-following ability~\citep{betker2023improving}.

\paragraph{Qualitative Results.}




We first visualize the zero-shot generation results on the MS-COCO validation set. In Figure~\ref{fig:coco_gen}, we compare the generated images of our method that takes blobs as input and GLIGEN that takes bounding boxes as input, where the ``ground-truth'' real images are also shown as a reference. 
We can see that our generated images are much more aligned with the reference images from both two perspectives: 1) visual appearance of each object, where the object color, shape and style have been better captured by the blob descriptions; and 2) its spatial arrangement, where the object pose and orientation have been better captured by blob parameters.
These results demonstrate that with blob representations, our method achieves much more fine-grained control over the generation. 

We then visualize various image editing results by manually changing a blob representation (\emph{e.g.}, either blob description or blob parameter) while keeping other blobs the same. See Appendix~\ref{sec:image_editing} for more details. In Figure~\ref{fig:coco_edit}, we show that our method can enable different local editing capabilities by solely changing the corresponding blob descriptions. In particular, we can change the object color, category, and texture while keeping the unedited regions mostly the same. Furthermore, our method can also make different object repositioning tasks easily achievable by solely manipulating the corresponding blob parameters. For instance, we can move an object to different locations, change an object's orientation, and add/remove an object while also keeping other regions nearly unchanged. Note that we have not applied any attention map guidance or sample blending trick~\citep{avrahami2022blended,ge2023expressive} during sampling, to preserve localization and disentanglement in the editing process. Thus, these results demonstrate that a well-disentangled and modular property naturally emerges in our blob-grounded generation framework.

\begin{table}[t]
\vspace{-6pt}
\setlength{\tabcolsep}{4.5pt}
\caption{Ablation studies on each component of our method separately. Here, to save computational cost, we train on 140K random samples and evaluate using 10K samples. All the models are evaluated after training for 150K steps with a batch size of 256.}
\label{tab:ablation}
\begin{center}
\begin{small}
\resizebox{0.9\columnwidth}{!}{
\begin{tabular}{lccccc}
\toprule
  & FID~$\downarrow$ & mIOU~$\uparrow$ & rCLIP$_t$~$\uparrow$  & rCLIP$_i$~$\uparrow$     \\
\midrule
Ours   & 9.14	& 0.4805	& 0.2859	& 0.8453      \\
\midrule
w/o blob emb. cat. & 9.21 &	0.4706	& 0.2849	& 0.8435       \\
w/o masking in CA	& 9.25	& 0.4401	& 0.2839 &	0.8410 \\
w/o Masked CA  & 9.47	& 0.4471	& 0.2840	& 0.8412       \\
w/o Gated SA  & 9.79	& 0.4898	& 0.2856	& 0.8422      \\
w/o prompt tuning  & 9.65	& 0.4430	& 0.2795	& 0.8248      \\
\bottomrule
\end{tabular}}
\end{small}
\end{center}
\vskip -0.1in
\end{table}

\subsubsection{Ablation Studies}
\label{sec:ablation}

In Table~\ref{tab:ablation}, we remove each component in our method separately and compare the resulting generation performance with our original design to highlight its impact. 

\paragraph{Blob Embedding Concatenation. } 
When blob text embeddings are directly passed to our Masked CA module, without concatenating blob parameter embeddings, we see a small performance reduction consistently across all four metrics. It implies adding blob embedding concatenation can slightly improve both generation quality and controllability. 

\paragraph{Masking in Masked CA. } When we remove masking in our Masked CA module, \emph{i.e.}, each blob text embedding does not attend only to visual features in the corresponding local region any more, we see a large performance drop on all three metrics: mIOU, rCLIP$_t$ and rCLIP$_i$, along with a slightly higher FID. It implies incorporating masking in Masked CA mainly improves the controllability. 

\paragraph{Masked CA. } When we remove the whole Masked CA module (where we only use the Gated SA module from GLIGEN to enable the blob control), we see a large performance drop consistently across all four metrics. It implies the importance of Masked CA in achieving both good generation quality and controllability. 

\paragraph{Gated SA. } Removing Gated SA results in a much worse FID but has a slightly mixed impact on other three metrics. On average, its impact on controllability is small. We hypothesize that it improves generation quality because it increases the expressive power of the base model. 

\paragraph{Prompt Tuning. } 
When we use a different prompt for LLaVA-1.5 to generate blob descriptions, which does not specifically ask the image captioning model to focus on the object itself (see Appendix~\ref{sec:prompt_tune_appendix} for details), we see a significant performance drop across all four metrics. It demonstrates the necessity of high data quality, in particular, a good blob description for each object.

\begin{table*}[t]
\caption{Evaluation of generation compositionality in terms of counting and spatial correctness on NSR-1K~\citep{feng2023layoutgpt}. Given an input prompt, our method uses LLMs to generate blob representations for blob-grounded image generation. For image accuracy, we use Grounding DINO~\citep{liu2023grounding} to detect bounding boxes from generated images.}
\label{tab:reasoning}
\begin{center}
\begin{small}
\resizebox{0.8\textwidth}{!}{
\begin{tabular}{lcccccc}
\toprule
\multirow{2}{*}{Method} & \multicolumn{4}{c}{Numerical Reasoning} & \multicolumn{2}{c}{Spatial Reasoning} \\ \cmidrule(lr){2-5} \cmidrule(lr){6-7} & Layout Prec & Layout Rec & Layout Acc & Image Acc & Layout Acc & Image Acc  \\
\midrule
\rowcolor{Gray}
\emph{Text $\to$ Image } & & & & & & \\ 
SD-1.4~\citep{rombach2022high} & - & - & - & 44.82 & - & 32.58 \\
SD-2.1~\citep{rombach2022high} & - & - & - & 48.49 & - & 32.20 \\
SDXL~\citep{podell2023sdxl} & - & - & - & 46.49 & - & 46.59 \\
Attend-and-Excite (SD-1.4)~\citep{chefer2023attend} &  - & - & - & 47.91 & - & 35.98 \\
Attend-and-Excite (SD-2.1)~\citep{chefer2023attend} &  - & - & - & 50.33 & - & 36.74 \\
\midrule
\rowcolor{Gray}
\emph{Text $\to$ Layout $\to$ Image } & & & & & & \\ 
LayoutGPT (GPT3.5-chat)~\citep{feng2023layoutgpt} & 75.40 & 86.23 & 74.62 & \underline{61.54} & 81.98 & 72.01 \\
LayoutGPT (GPT4)~\citep{feng2023layoutgpt} & \textbf{81.02} & 85.63 & \underline{78.11} & 60.25 & \underline{86.23} & 74.35 \\
Ours (GPT3.5-chat) & \underline{76.08} & \underline{86.49} & 75.75 & 60.46 & 83.27 & \underline{75.83} \\
Ours (GPT4) & 75.73 & \textbf{86.77} & \textbf{78.67} & \textbf{62.96} & \textbf{90.23} & \textbf{80.16} \\
\bottomrule
\end{tabular}}
\end{small}
\end{center}
\end{table*}

\begin{figure*}[t]
    \centering
    \includegraphics[width=0.97\textwidth]{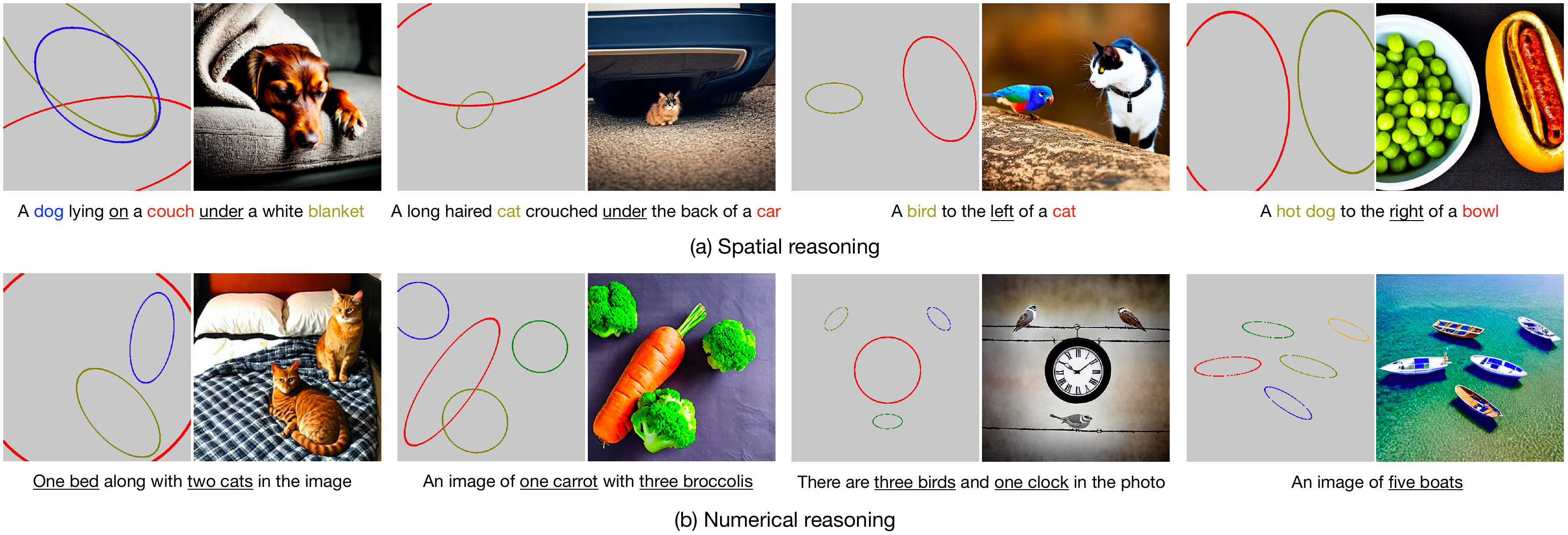}
    \vskip -0.15in
    \caption{Qualitative results of our method on two compositional generation tasks of NSR-1K~\citep{feng2023layoutgpt}: (a) spatial reasoning and (b) numerical reasoning. Given a caption, we prompt GPT4 to generate blob parameters (Left) and LLAMA-13B to generate blob descriptions (not shown in the figure), which are passed to our blob-grounded text-to-image model to synthesize an image (Right). }
    \label{fig:nsr_reasoning}
\end{figure*}

\subsection{LLMs for Blob Generation}

In this section, we compare our method with previous approaches on compositional reasoning, by prompting LLMs to generate blob representations. We closely follow the evaluation protocol proposed by LayoutGPT~\citep{feng2023layoutgpt} for the comparison. We defer a comparison with other LLM-grounded generation methods (\emph{i.e.,} LMD~\citep{lian2023llmgrounded}) in a different setting to Appendix~\ref{sec:wo_retrieval}.

\subsubsection{Experiment Setup}

\paragraph{Data Preparation.}
The original NSR-1K benchmark proposed by~\citep{feng2023layoutgpt} is mainly targeted for models that use bounding boxes as grounding input. To evaluate our method on NSR-1K, we need to replace bounding boxes with blob representations. Specifically, since almost all images in NSR-1K come from MS-COCO, we can convert their ground-truth segmentation maps to the blob parameters using the same ellipse fitting optimization algorithm, and we use LLaVA-1.5 to generate blob descriptions from cropped images. This results in 738 training and 264 testing examples\footnote{Note that 19 images in the original test set are not from MS-COCO so we do not include them for evaluation.} for spatial reasoning, and 38,698 training and 762 testing examples for numerical reasoning.

\paragraph{Evaluation Metrics.}
Similar to LayoutGPT~\citep{feng2023layoutgpt} that evaluates the layout planning performance of LLMs, we report precision (Prec), recall (Rec), and accuracy (Acc) based on generated layout counts and their spatial positions. To evaluate layout consistency from generated images, we observed that the detection model GLIP~\citep{li2022grounded}, which was employed by LayoutGPT, frequently misses salient objects (see Appendix~\ref{sec:glip_appendix} for details). Thus, we use the more recently proposed detector Grounding DINO~\citep{liu2023grounding} to obtain more accurate bounding boxes. We then compute the average accuracy based on the detected bounding boxes and the ground-truth ones by following the same evaluation pipeline.

\subsubsection{Numerical and Spatial Reasoning }

\paragraph{Quantitative Results.}
In Table~\ref{tab:reasoning}, we show the performance of different models on both numerical reasoning and spatial reasoning. For the layout planning evaluation, we observe that both GPT3.5-chat and GPT4 can generate good blob parameters with our customized in-context learning procedure. For instance, the layout accuracies of GPT4 are 78.67\% versus 78.11\% on counting correctness and 90.23\% versus 86.23\% on spatial correctness for our method and LayoutGPT, respectively. For image evaluation, our method also achieves consistently better accuracies than LayoutGPT in both two tasks (counting: 62.96\% versus 61.54\%, spatial: 80.16\% versus 74.35\%). 
Compared with text-to-image models that do not incorporate layout planning, notably SDXL, our results distinctly highlight the significance of our blob-grounded framework in facilitating more reliable generation with better prompt following capabilities. 

\vspace{-5pt}
\paragraph{Qualitative Results.}

In Figure~\ref{fig:nsr_reasoning}, we visualize the blobs generated by GPT4 and images generated by our blob-grounded model for various prompts from both spatial and numerical reasoning tasks. We defer more visualization results of a side-by-side comparison among different models to Appendix~\ref{sec:comp_reason_supp}.
We can see that our method can not only synthesize images that align well with the layouts generated by GPT4, which supports our quantitative results, but can also achieve high photo-realism. In particular, we observe that the out-of-context ``crop-and-paste'' effect less frequently appears in our generated images. Instead, objects have been rendered in a coherent and natural way, along with an appropriate background, to follow physical laws and visual commonsenses. For instance, three birds are standing on electric wires (b, third example) and the cat is curiously looking at a colorful bird on the left (a, third example). 



\vspace{-5pt}
\section{Conclusions}
\label{concl}
\vspace{-2pt}

In this work, we proposed to ground existing text-to-image generative models on blob representations for compositional generation. In particular, we applied the open-vocabulary segmentation and vision-language models to extract blob parameters and blob descriptions, which contain rich spatial and semantic information of images. We then introduced a blob-grounded generative model, termed \sname, where a new masked cross-attention module that takes blobs as grounding inputs is injected into pre-trained text-to-image models. Furthermore, to leverage the compositional ability of LLMs for image generation, we designed a new in-context learning approach for LLMs to infer blob representations from text prompts. 
Finally, we performed extensive experiments to show the superior performance of our method in various generative tasks, including layout-guided generation, image editing and compositional reasoning.

\vspace{-7pt}
\paragraph{Limitations.} Our work has several limitations that we leave for the future work. First, even though blob representations can preserve fine-grained details of the image, we cannot solely rely on them to perfectly recover the original image, where a combination with inversion methods~\citep{mokady2023null} is still needed. Second, we see some failure cases for image editing (see Figure~\ref{fig:coco_edit_fail_supp}), which we believe advanced editing techniques~\citep{avrahami2022blended} can be applied to alleviate. Third, we also see some failure cases in the numerical and spatial reasoning tasks (see Figure~\ref{fig:coco_reason_fail_supp}). It is an interesting challenge to further improve the integration between LLMs and blob-grounded generation. 

\section*{Impact Statement}





Our blob-grounded text-to-image model is based on a pre-trained text-to-image model, so it may inherit the potential biases and malicious information from the base model. Because our approach improves both the generation quality and the user controllability over the base model, on the positive side, it will improve the efficiency of human users in using generative models for creative work; on the negative side, similar to any generative AI tool, it can be used to generate malicious content. 
Furthermore, when augmented by LLMs for layout planning, our method will simplify the layout designing process, resulting in less burden on human designers for content creation. But we also note that the biases and misinformation from LLMs could also harm the use of our approach without proper regulation.


\bibliography{references}
\bibliographystyle{icml2024}

\newpage
\appendix
\onecolumn

\section{Implementation Details}


\subsection{Synthetic Global Captions}
\label{sec:syn_caption_appendix}

\begin{figure}[h]
    \centering
    \includegraphics[width=\textwidth]{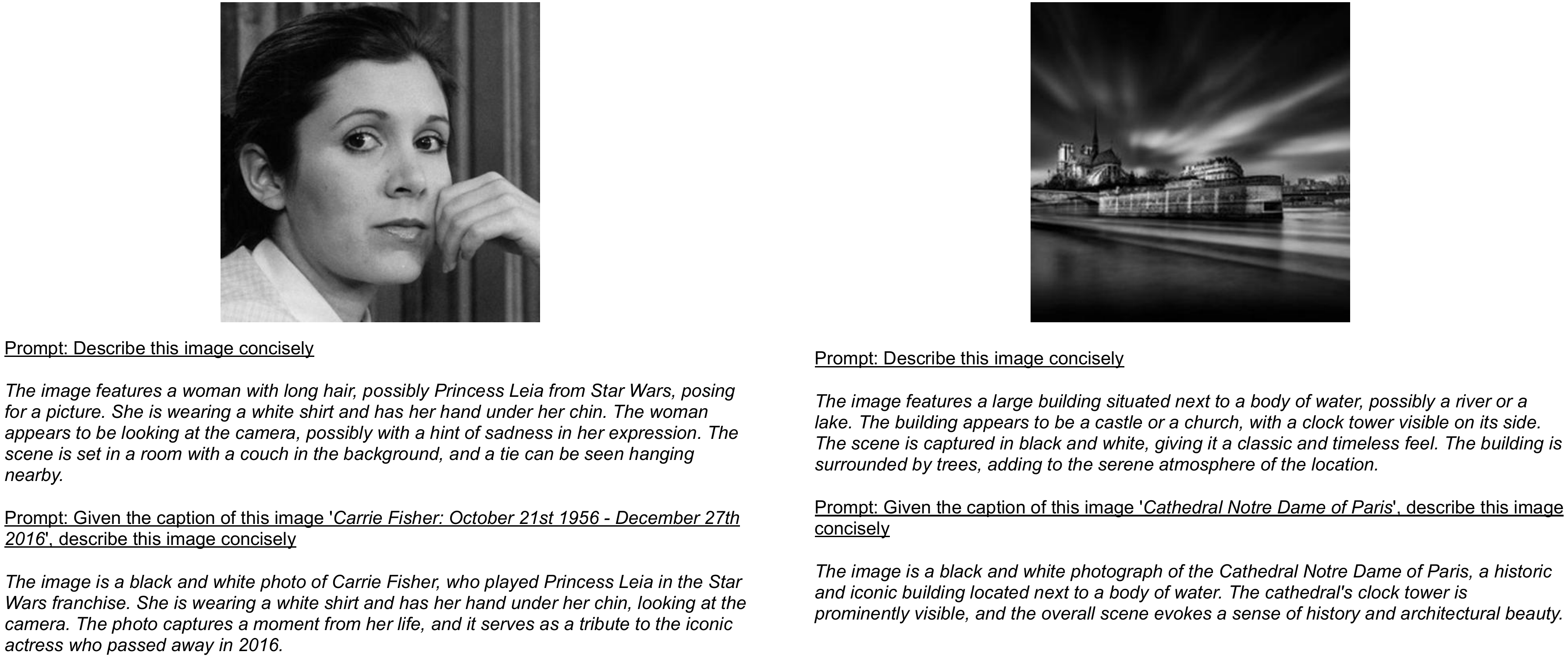}
    \vskip -0.1in
    \caption{Comparison between two prompts for LLaVA-1.5 to generate synthetic global captions, where we show the response of LLaVA-1.5 for each prompt on an image. We can see that incorporating the original real caption into the prompt can capture the specialized information that LLaVA-1.5 cannot provide.}
    \label{fig:global_cap_supp}
\end{figure}

To train our model, we use synthetic global captions instead of the noisy real captions from the original image-text dataset, as evidenced by recent findings that high-quality synthetic captions can improve the model's prompt following performance~\citep{betker2023improving}. To generate synthetic captions, we use LLaVA-1.5 as the image captioner with the prompt: \texttt{"Given the caption of this image '<real caption>', describe this image concisely."}. Compared with the standard prompt: \texttt{"Describe this image concisely."}, it can capture the regional information that LLaVA-1.5 cannot provide to incorporate the real caption into the prompt, as shown in Figure~\ref{fig:global_cap_supp}. 

\subsection{Prompt Tuning for Blob Descriptions}
\label{sec:prompt_tune_appendix}

When we use the image captioner (\emph{e.g.}, LLaVA-1.5) to describe the cropped image as depicted by the blob parameter, we need to make the blob description to capture fine-grained visual features of the local region. To this end, we consider two types of prompts: one is to explicitly ask for visual features, and the other one is to provide the context of what the full image is about. 
Specifically, we compare the following two prompts: 
\begin{itemize}[leftmargin=*]\setlength\itemsep{-0em}
    \item \texttt{"Can you briefly describe this <category name> in the close-up and focus on its color, appearance, size, and style, etc.?"}
    \item \texttt{"While in the original full image, '<global caption>', can you briefly describe this <category name> in the close-up?"}
\end{itemize}
Note that we assume the category name for the object in the local region is given, which can come as an output of an off-the-shelf image segmentation model. 
As evidenced by the matrix scores in Table~\ref{tab:ablation}, the first prompt works better than the second one. Therefore, we use the first prompt to generate blob descriptions for all our experiments.

\subsection{In-Context Learning for LLMs}

We introduce a new in-context learning approach that contains two separate procedures to generate blob parameters and blob descriptions, respectively. 

\subsubsection{Blob Parameter Generation}
\label{sec:blob_param_appendix}

We mainly use GPT3.5-chat~\citep{ouyang2022training} and GPT4~\citep{achiam2023gpt} to generate blob parameters. To make GPT3.5-chat/GPT4 better understand and infer the numerical values, we follow \cite{feng2023layoutgpt} to use the CSS format to represent blob parameters. The final prompt for GPT3.5-chat/GPT4 consists of a system prompt that instructs the blob parameter generation, $k$ demonstration examples, and the test prompt (usually a global caption). In Table~\ref{tab:prompt_param}, we use \texttt{"a teddy bear to the left of a bed"} as the text prompt of a real example to show the system prompt.

\begin{table}[h]
    \centering
    \footnotesize
    \caption{The system prompt for GPT3.5-chat/GPT4 to generate blob parameters for the example \texttt{"a teddy bear to the left of a bed"}. Note that we have ignored the wrapped prompts specifically for GPT3.5-chat/GPT4, including \texttt{"role: system, content: "}, \texttt{"role: user, content: "} and \texttt{"role: assistant, content: "} for the sake of readability. }
    \vspace{6pt}
    \begin{tabular}{>{\RaggedRight}p{17cm}}
        \toprule
         Instruction: Given a sentence prompt that will be used to generate an image, plan the layout of the image. The generated layout should follow the CSS style, where each line starts with the object name and is followed by its absolute position depicted as an ellipse. Formally, each line should be like "object {\{major-radius: ?px; minor-radius: ?px; cx: ?px; cy: ?px; angle: ?\}}". The image is 512px wide and 512px high. Therefore, all properties of the positions (including major-radius, minor-radius, cx and cy) should not exceed 512px, and the value of angle is in degree and it should be within [0, 180]. Finally, we prefer all objects to be large (i.e., each ellipse better has a large major-radius), if possible. \\
         \\
         Prompt: a teddy bear to the right of a cat
         \\
         Layout: \\
         teddy-bear \{major-radius: 162px; minor-radius: 76px; cx: 444px; cy: 258px; angle: 96\} \\
         cat \{major-radius: 137px; minor-radius: 116px; cx: 149px; 236cy: ?px; angle: 3\} \\
         \\
         $[$ADDITIONAL $k-1$ DEMONSTRATION EXAMPLES REMOVED FOR SIMPLICITY$]$\\
         \\
         Prompt: a teddy bear to the left of a bed
         \\
         Layout:
         \\
         \bottomrule
    \end{tabular}
    \label{tab:prompt_param}
    \vspace{-5pt}
\end{table}

\subsubsection{Blob Description Generation}
\label{sec:blob_desc_appendix}

We mainly use LLaMA-13B to generate blob descriptions. Because the blob descriptions are just a list of text sentences, we use the simple text format for its generation, where the category name of an object is still used as a blob separator. The final prompt for LLaMA-13B includes a system prompt that instructs the blob description generation, $k$ demonstration examples, and the test prompt (usually a global caption). In Table~\ref{tab:prompt_desc}, we also use \texttt{"a teddy bear to the left of a bed"} as a real example to show the system prompt.

\begin{table}[h]
    \centering
    \footnotesize
    \caption{The system prompt for LLaMA-13B to generate blob descriptions for the example \texttt{"a teddy bear to the left of a bed"}. }
    \vspace{6pt}
    \begin{tabular}{>{\RaggedRight}p{17cm}}
        \toprule
         Instruction: Given a sentence prompt that will be used to generate an image, plan the region descriptions of the image, where each line starts with the object name. For example, each line should be like "cat \{The cat in the close-up is a large, gray and white cat with a fluffy appearance. The cat's size and style suggest that it is a domesticated cat, likely a house cat, and it is comfortable in its environment. The cat's gray and white coloration adds to its unique and visually appealing appearance.\}". The generated region description should describe the object in the close-up and focus on its color, appearance, size, and style, etc. \\
         \\
         Prompt: a teddy bear to the right of a cat
         \\
         Region Desc: \\
         teddy-bear \{The teddy bear in the close-up is white and has a large size. It is sitting next to a pink stuffed animal, which appears to be a dragon or a panda. The teddy bear is positioned on a bed, and it is surrounded by other stuffed animals, creating a cozy and playful scene.\} \\
         cat \{The cat in the close-up is a large, striped tabby cat. It has a distinctive black and brown striped pattern on its fur, which is quite noticeable. The cat appears to be sitting or standing on top of a stuffed animal, possibly a teddy bear, which adds a playful and curious element to the scene. The cat's size and style give it a unique and eye-catching appearance, making it an interesting subject for a close-up photo.\} \\
         \\
         $[$ADDITIONAL $k-1$ DEMONSTRATION EXAMPLES REMOVED FOR SIMPLICITY$]$\\
         \\
         Prompt: a teddy bear to the left of a bed
         \\
         Region Desc:
         \\
         \bottomrule
    \end{tabular}
    \label{tab:prompt_desc}
\end{table}

\subsection{Examples of Blob Representations}

\begin{figure}[H]
    \centering
    \includegraphics[width=\textwidth]{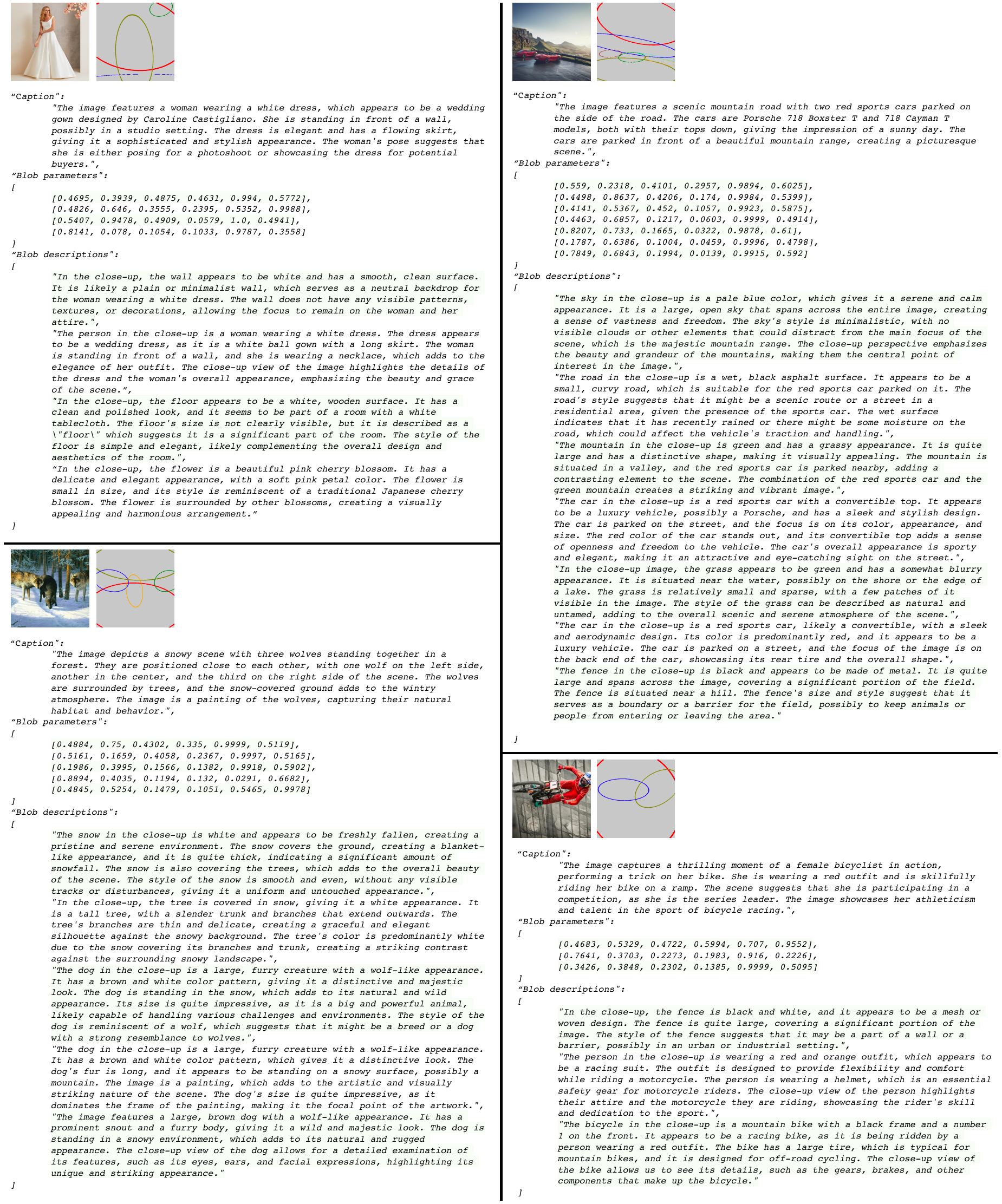}
    \vskip -0.1in
    \caption{Four examples of decomposing a scene into blob representations, where we visualize the blob ellipses by the side of each image, and also include the synthetic global captions that are used to train our model. All blob parameters are normalized to the range of [0, 1]. }
    \label{fig:blob_exam_supp}
\end{figure}

\subsection{Unreliability of GLIP}
\label{sec:glip_appendix}

\begin{figure}[H]
    \centering
    \includegraphics[width=\textwidth]{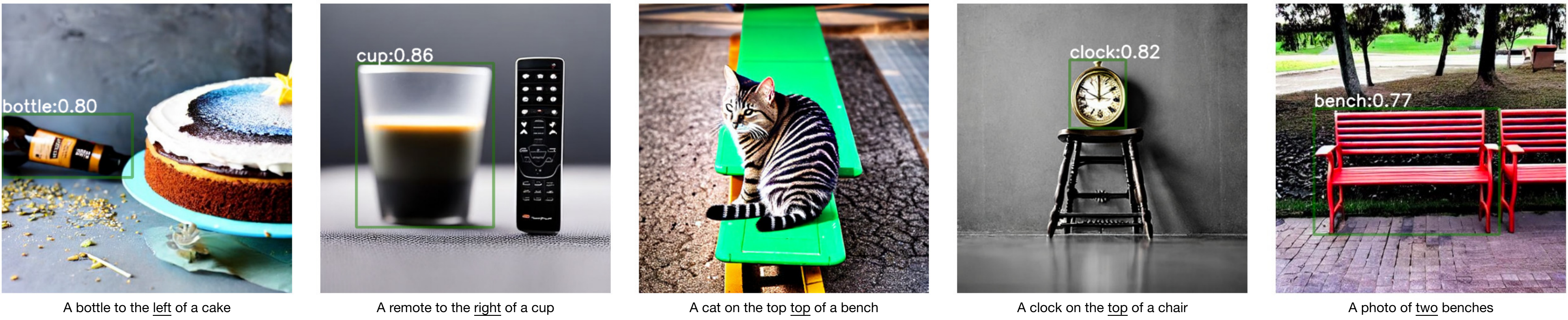}
    \vskip -0.1in
    \caption{A few examples that GLIP fails to detect all correctly generated objects from the input prompts. All detected objects have been marked a bounding box with a prediction score. }
    \label{fig:glip_supp}
\end{figure}

LayoutGPT has used GLIP~\citep{li2022grounded} to detect objects from generated images for evaluating the spatial and numerical correctness of the generation. However, we found that GLIP could consistently miss objects in a generated image, as shown in Figure~\ref{fig:glip_supp}. As a result, we use a more recently developed detection method, called Grounding DINO~\citep{liu2023grounding}, as a better proxy for our evaluation.

\subsection{Detailed Process of Image Editing}
\label{sec:image_editing}
Given an image generated from \sname and its conditioning information (including blob representations, global text, and initial Gaussian noise), we only change its blob representations and then pass the new blob representations along with the original global text and the original initial Gaussian noise to \sname to get the edited image. Note that we follow the standard denoising sampling to generate the edited image without relying on any advanced editing technique (i.e., attention guidance or image blending), implying good disentanglement of \sname.

When we change a blob representation (blob parameter and blob description), we can either change the blob parameter (defined as $\tau = [c_x, c_y, a, b, \theta]$) for repositioning, rotating and moving the object, or change the blob description (defined as an object-level text caption) for editing the object’s appearance and other visual features. Moreover, we can completely remove/add a blob representation to remove/add the object.

\section{Additional Qualitative Results}

\subsection{Blob Representations Capture Irregular, Large Objects and Background}

\begin{figure}[t]
    \centering
    \includegraphics[width=\textwidth]{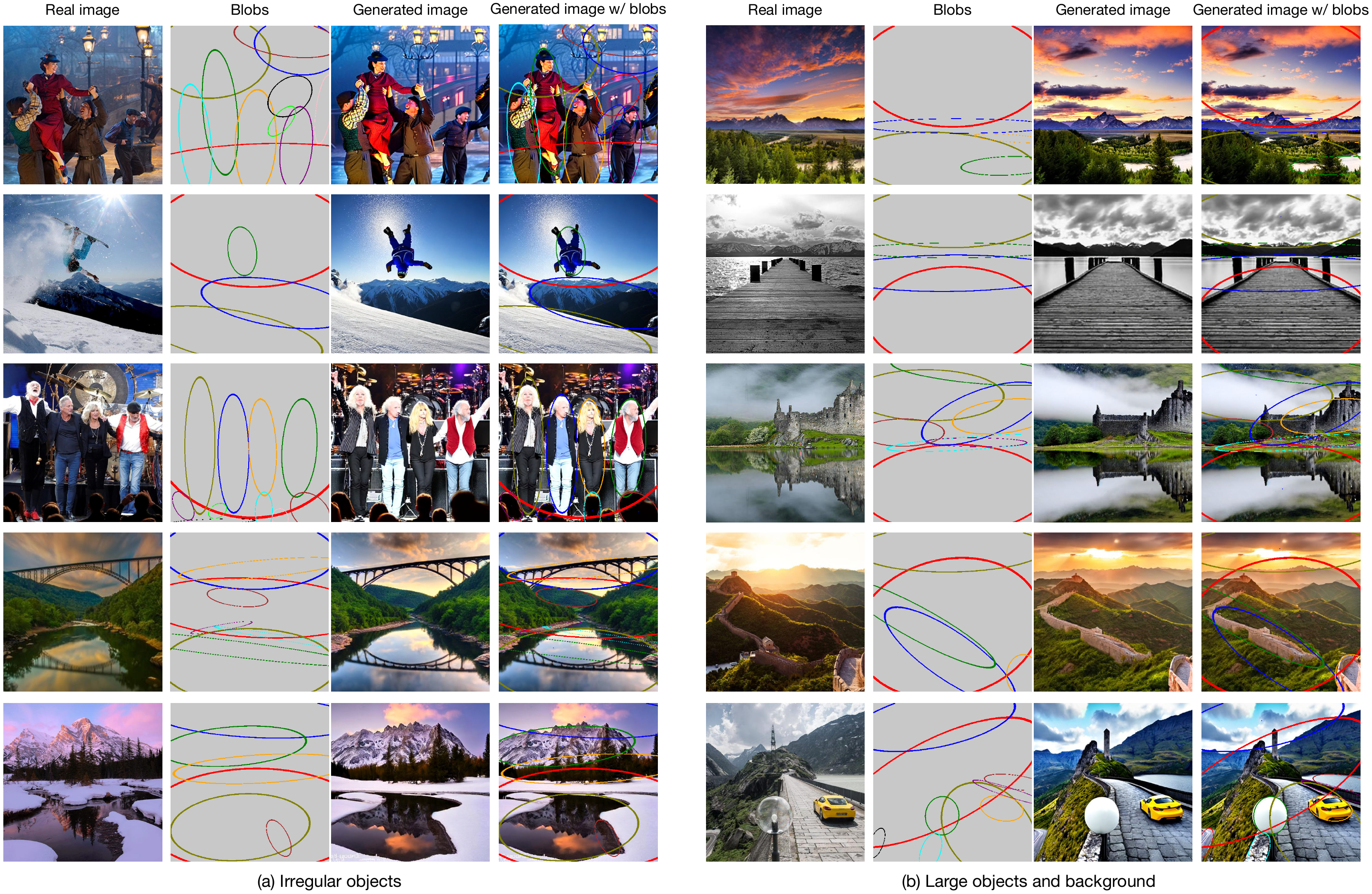}
    \vskip -0.1in
    \caption{Examples of blob-grounded generation capturing (a) irregular objects (e.g. ``person with waving hands" and ``river"), and (b) large objects and background (e.g., ``sky", ``mountain" and ``grass"). From left to right in each row, we show the reference real image, blobs, generated image and generated image superposed on blobs. Better zoom in for better visualization. }
    \label{fig:blobgen_irregular_large}
\end{figure}

To show blobs can capture irregular, large objects and background, we provide extra qualitative results in Figure~\ref{fig:blobgen_irregular_large}. 

\paragraph{Irregular Objects.} 
Examining Figure~\ref{fig:blobgen_irregular_large}(a), the first three rows show that blob representations can capture ``a person with waving hands outside the blob''. Note that “person” blobs allow the person’s hands to be outside their ellipse region and still capture their pose accurately. The last two rows show that blob representations can capture ``a river with irregular shapes". Moreover, the fourth row in Figure~\ref{fig:blobgen_irregular_large}(b) shows that blob representations can capture ``the great wall with a zigzag shape". 

Two factors allow our blob representation to capture irregular shapes: 1) Training data contains many irregular objects where our blobs are designed to allow some parts of the irregular object to be outside the blob ellipse. Thus, the trained model can quickly learn this design from the training data. 2) More fine-grained shape details of the irregular objects can be captured by the text sentences of blob descriptions  (e.g., a blob description may contain “a zigzag river” or “an upside-down person in the air with two arms widely open”), which complement the spatial depiction of blob parameters. 3) In some cases with very large irregular objects, such as ``the great wall with a zigzag shape", multiple blobs can be used to capture each individual part of the object, or neighboring objects (e.g. sand, rock, etc) can help with creating a particular irregular shape.

\paragraph{Large Objects and Background.} 
Examining Figure~\ref{fig:blobgen_irregular_large}(b), the first two rows show that blob representations can capture the large ``sky" of a similar color and mood to that in the reference real image. The second row shows that blob representations can capture the large ``pier" of a similar color, pattern, and shape to that in the reference real image. The same applies to the ``foggy grass" and its reflection in the water in the third row and the large mountains in the last two rows.

The rationale behind this capability is that: 1) We can always use as large blob ellipses as possible to fit the large objects, and we do not necessarily restrict the whole blob ellipse area to be within the pixel range. This provides more flexibility for blobs to capture extremely large objects and backgrounds. 2) More importantly, the text sentence description from the blob representation can effectively help capture the fine-grained details (i.e., color, appearance, texture, etc.) of the large object.

\subsection{Zero-Shot Generation on MS-COCO}

\begin{figure}[t]
    \centering
    \includegraphics[width=0.75\textwidth]{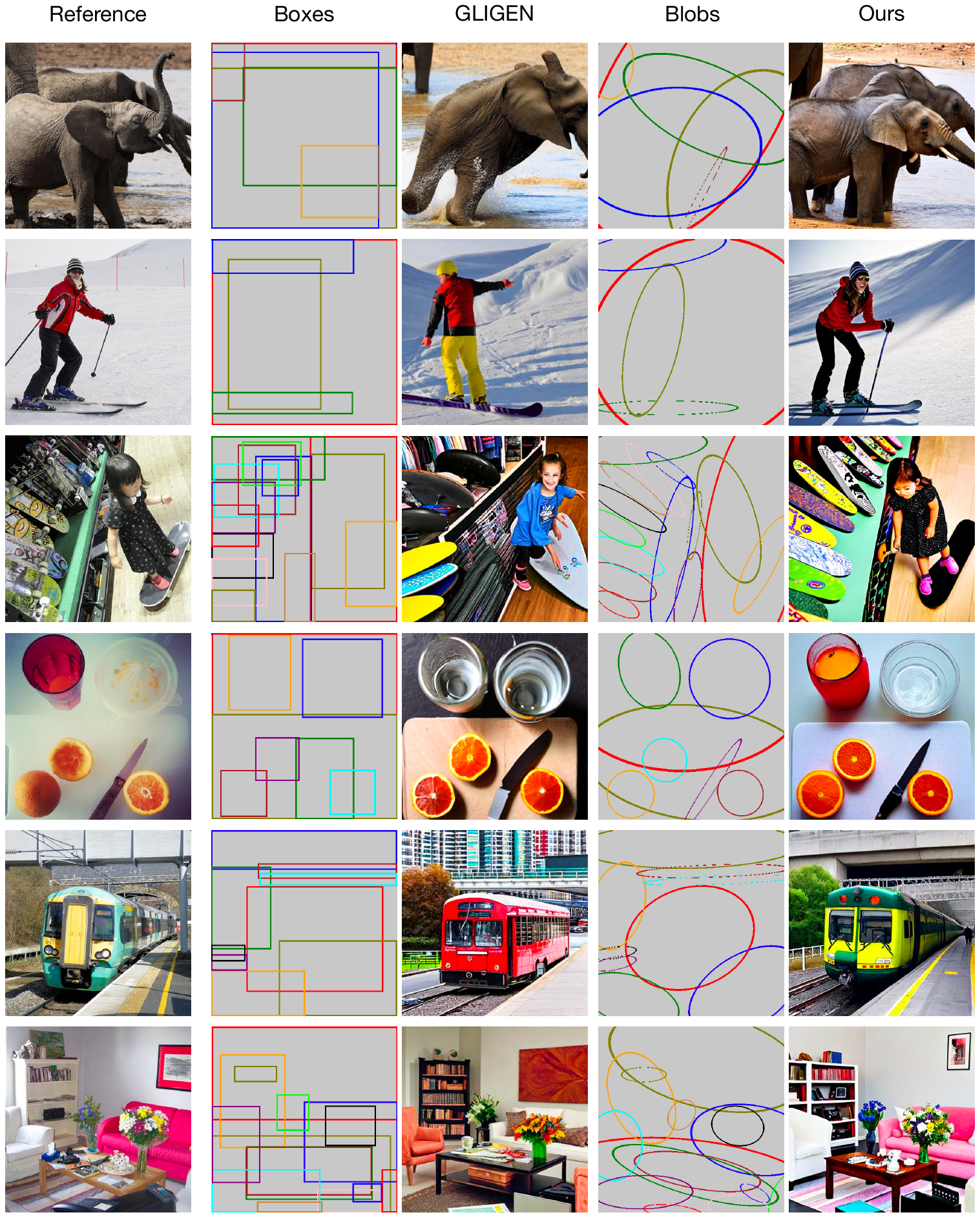}
    \vskip -0.1in
    \caption{Zero-shot layout-grounded generation results of GLIGEN and our method on MS-COCO validation set. In each row, we visualize the reference real image (Left), bounding boxes and GLIGEN generated image (Middle), blobs and our generated image (Right). All images are in resolution of 512$\times$512. }
    \label{fig:coco_gen_supp}
\end{figure}

We show more zero-shot layout-grounded generation results of GLIGEN and our method in Figure~\ref{fig:coco_gen_supp}. In general, our method can capture more fine-grained details of the original model, using dense blob representations.

\subsection{Image Editing on MS-COCO}

\begin{figure}[t]
    \centering
    \includegraphics[width=\textwidth]{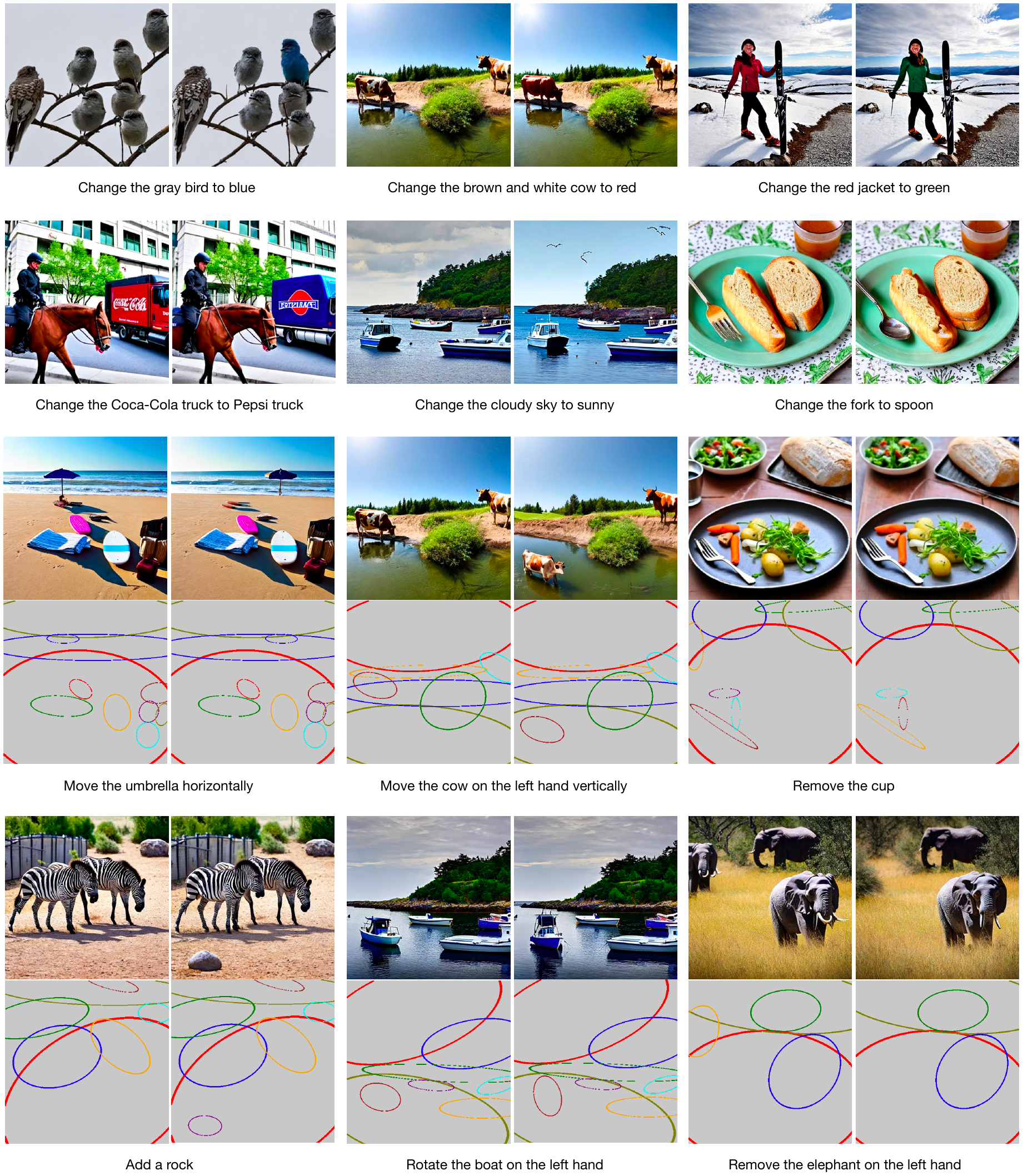}
    \vskip -0.1in
    \caption{Various image editing results of our method on the MS-COCO validation set, where each example contains two generated images: (Left) original setting and (Right) after editing. The top two rows show the local editing results where we only change the blob description and since the blob parameters stay the same after editing, we do not show blob visualizations. The bottom four rows show the object reposition results where we only change the blob parameter. All images are in resolution of 512$\times$512. }
    \label{fig:coco_edit_supp}
\end{figure}

We show more image editing results of our method in Figure~\ref{fig:coco_edit_supp}. Note that we perform various image editing tasks by merely modifying the blob parameter for object repositioning, or modifying the blob description for local object/attribute manipulation. We do not use any advanced image editing technique, such as attention mask guidance~\citep{ge2023expressive} or sample blending~\citep{avrahami2022blended}, to maintain localization and disentanglement for editing. As we can see, our method can enable various image editing capabilities, including changing the fine-grained orientation of an object that previous box layouts can hardly work out. With these promising editing results, we believe our blob-grounded generative model in general has a well-disentangled property, with a modular control over generation for each local region depicted by blob representations.

\subsection{Numerical and Spatial Reasoning}
\label{sec:comp_reason_supp}

\begin{figure*}[t]
    \centering
    \includegraphics[width=\textwidth]{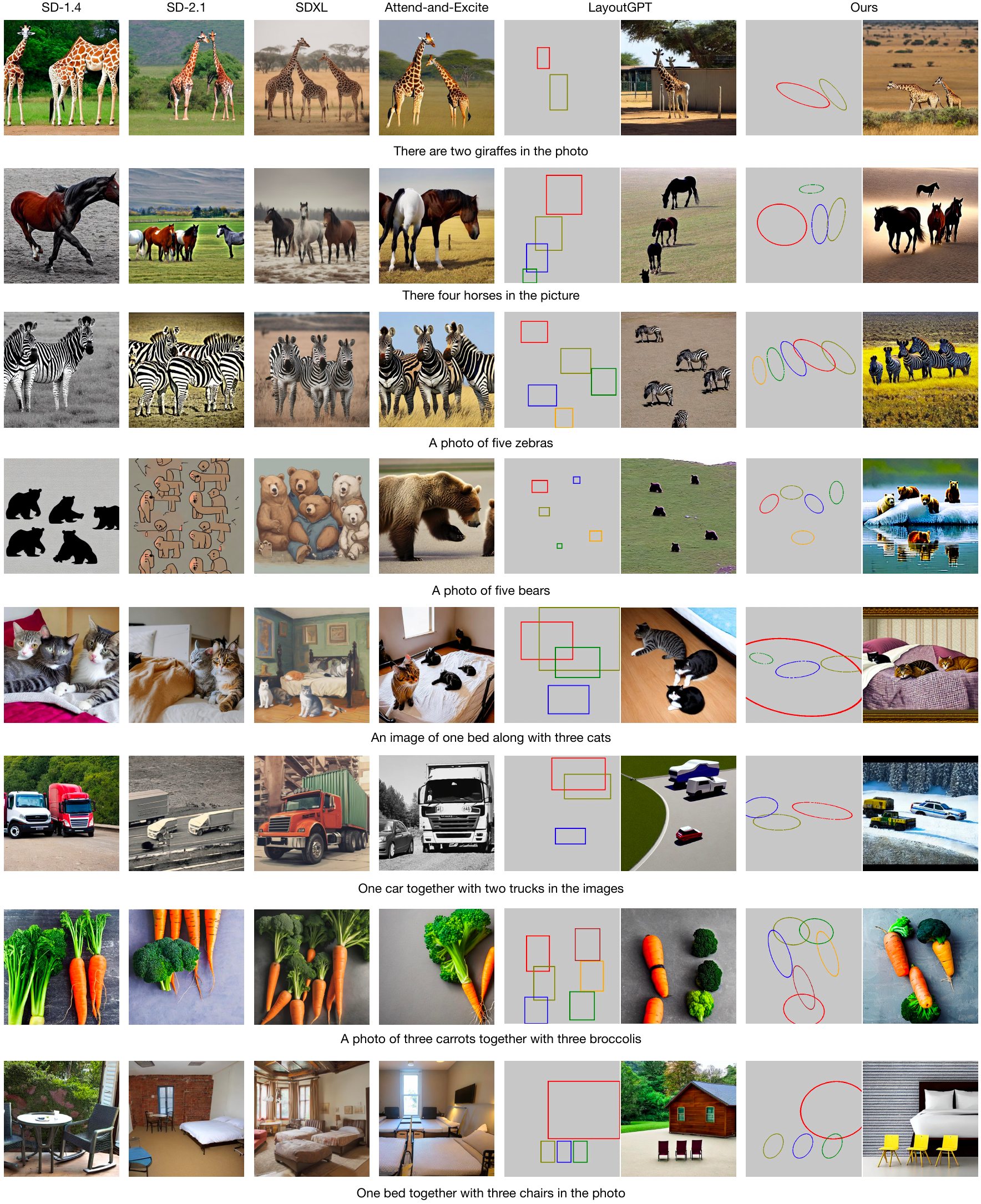}
    \vskip -0.1in
    \caption{Qualitative results of various methods on numerical reasoning tasks. In our method, given a caption, we prompt GPT4 to generate blob parameters (Left) and LLAMA-13B to generate blob descriptions (not shown in the figure), which are passed to our blob-grounded text-to-image generative model to synthesize an image (Right).}
    \label{fig:nsr_counting_supp}
\end{figure*}

\begin{figure*}[t]
    \centering
    \includegraphics[width=\textwidth]{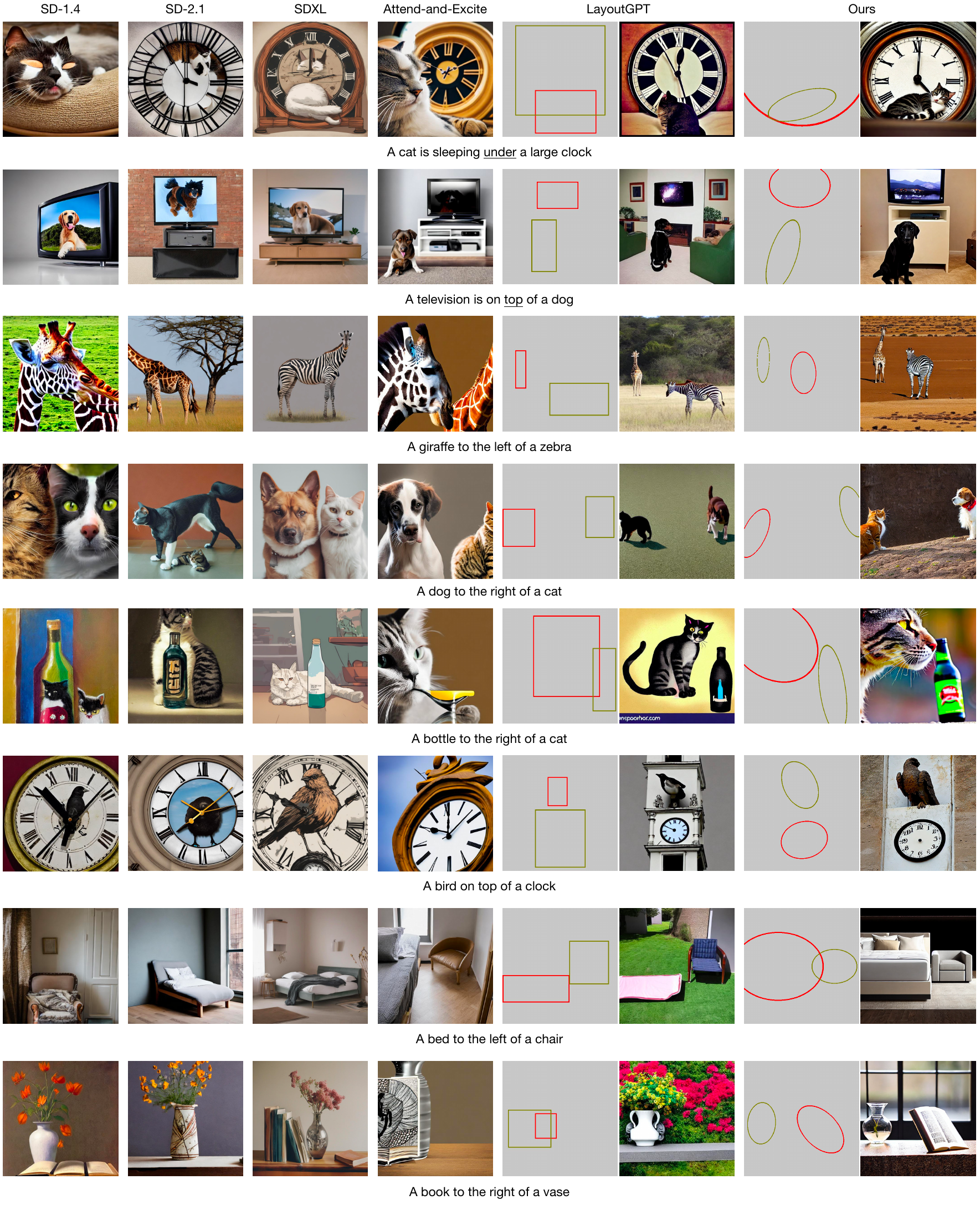}
    \vskip -0.1in
    \caption{Qualitative results of various methods on spatial reasoning tasks. In our method, given a caption, we prompt GPT4 to generate blob parameters (Left) and LLAMA-13B to generate blob descriptions (not shown in the figure), which are passed to our blob-grounded text-to-image generative model to synthesize an image (Right).}
    \label{fig:nsr_spatial_supp}
\end{figure*}

We show more numerical and spatial reasoning results of our method and previous approaches in Figures~\ref{fig:nsr_counting_supp} and \ref{fig:nsr_spatial_supp}, respectively. In addition, we visualize the layouts inferred by GPT4 for both GLIGEN and our method. We can see that all methods without layout planning in the middle always fail in the task, including SDXL~\citep{podell2023sdxl} that has a large model size and more advanced training procedures, and Attention-and-Excite~\citep{chefer2023attend} that has an explicit attention guidance. Compared with LayoutGPT based on GLIGEN, our method can not only generate images with better spatial and numerical correctness, but also in general has better visual quality with less ``copy-and-paste'' effect.

\subsection{Using Fixed In-context Examples without Retrieval}
\label{sec:wo_retrieval}

\subsubsection{Comparison between with and without Retrieval}

Here, we show retrieval from a large blob dataset for getting in-context demo examples is \textit{not necessary} for our method. The main reason why we use retrieval as a demonstration is that we want to follow the exact evaluation protocol of LayoutGPT to make a fair comparison with LayoutGPT on the NSR-1K benchmark. To make a direct comparison between using fixed in-context examples (“fixed”) vs. using retrieved in-context examples (“retrieval”), we summarize the results in Table~\ref{tab:comp_fixed_retrieval}. We can see that although using the fixed in-context examples underperforms using retrieved in-context examples in both spatial and numerical reasoning tasks, the performance gap is not large. It implies the effectiveness of our method with the use of fixed in-context examples.

\begin{table*}[h]
\caption{Comparison between using fixed in-context examples (“fixed”) vs. using retrieved in-context examples (“retrieval”) for our method in terms of counting and spatial correctness on NSR-1K~\citep{feng2023layoutgpt}.}
\label{tab:comp_fixed_retrieval}
\begin{center}
\begin{small}
\begin{tabular}{lcccccc}
\toprule
\multirow{2}{*}{Method} & \multicolumn{4}{c}{Numerical Reasoning} & \multicolumn{2}{c}{Spatial Reasoning} \\ \cmidrule(lr){2-5} \cmidrule(lr){6-7} & Layout Prec & Layout Rec & Layout Acc & Image Acc & Layout Acc & Image Acc  \\
\midrule
retrieval & \textbf{75.73}    &    \textbf{86.77}    &   \textbf{78.67} &  \textbf{62.96} & 90.23    &    \textbf{80.16} \\
fixed & 71.91   &    86.07   &     77.95   &    60.74 &  \textbf{90.80}   &    77.84 \\
\bottomrule
\end{tabular}
\end{small}
\end{center}
\vskip -0.1in
\end{table*}

\subsubsection{Comparison between Our Method and LMD~\citep{lian2023llmgrounded}}

In Table~\ref{tab:comp_lmd}, we show when we use the same 8 fixed in-context demo examples (without retrieval) to prompt GPT-4 for blob generation, our method outperforms LMD~\citep{lian2023llmgrounded}, a strong baseline that has used fixed in-context examples for prompting LLMs to generate bounding boxes.

\begin{table*}[h]
\caption{Comparison between our method and LMD~\citep{lian2023llmgrounded} in terms of counting and spatial correctness on NSR-1K~\citep{feng2023layoutgpt}, where both two methods use the same 8 fixed in-context demo examples to prompt GPT4 for layout generation.}
\label{tab:comp_lmd}
\begin{center}
\begin{small}
\begin{tabular}{lcccccc}
\toprule
\multirow{2}{*}{Method} & \multicolumn{4}{c}{Numerical Reasoning} & \multicolumn{2}{c}{Spatial Reasoning} \\ \cmidrule(lr){2-5} \cmidrule(lr){6-7} & Layout Prec & Layout Rec & Layout Acc & Image Acc & Layout Acc & Image Acc  \\
\midrule
LMD~\citep{lian2023llmgrounded} & 71.76    &    85.96    & \textbf{78.02} & 57.61 &  83.86  &   73.50 \\
Ours & \textbf{71.91}   &   \textbf{86.07}   &     77.95   &    \textbf{60.74} &  \textbf{90.80}   &    \textbf{77.84} \\
\bottomrule
\end{tabular}
\end{small}
\end{center}
\vskip -0.1in
\end{table*}

\begin{figure}[t]
    \centering
    \includegraphics[width=\textwidth]{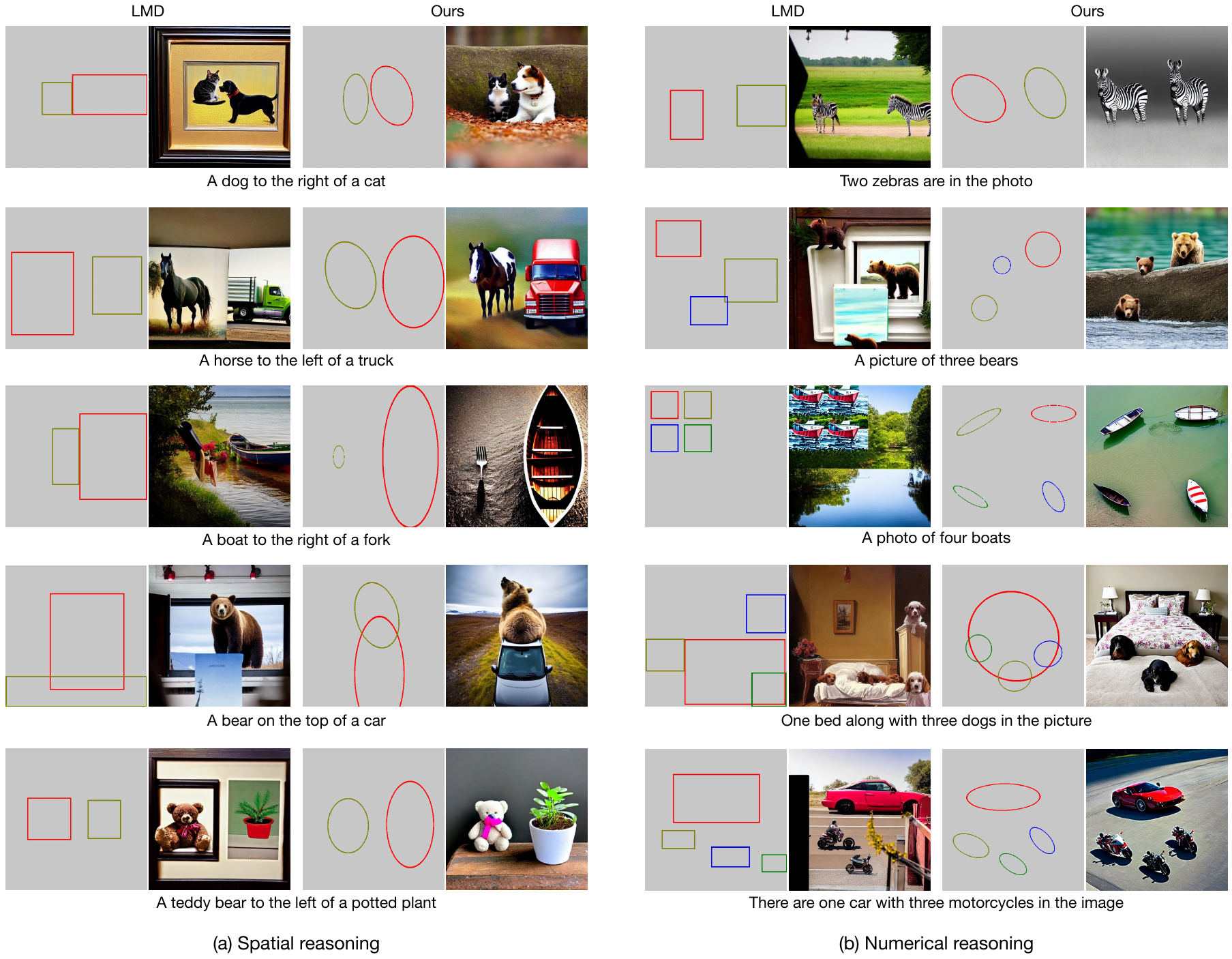}
    \vskip -0.1in
    \caption{Qualitative results of comparing our method with LMD~\citep{lian2023llmgrounded} on the NSR-1K benchmark for spatial and numerical reasoning. In each example, we first prompt GPT4 to generate boxes for LMD and blobs for our method, respectively, with the same 8 fixed in-context demo examples. The generated boxes and blobs are passed to LMD and BlobGEN to generate images, respectively. Better zoom in for better visualization. }
    \label{fig:blobgen_comp_lmd}
\end{figure}

We also show the qualitative results of comparing our method and LMD in Figure~\ref{fig:blobgen_comp_lmd}. We observe that: 1) In some complex examples, such as ``a boat to the right of a fork" in Figure~\ref{fig:blobgen_comp_lmd}(a) and ``there are one car with three motorcycles in the image" in Figure~\ref{fig:blobgen_comp_lmd}(b), LMD fails but our method works. It confirms the quantitative results in Table~\ref{tab:comp_lmd}. 2) LMD consistently has the "copy-and-paste" artifacts in its generated images in which objects are put together without matching their context (since it modifies the diffusion sampling process to enforce compositionality, which may largely deviate from the original data denoising trajectory), such as the example of ``a teddy bear to the left of a potted plant" in Figure~\ref{fig:blobgen_comp_lmd}(a) and the example of ``a photo of four boats'' in Figure~\ref{fig:blobgen_comp_lmd}(b). In contrast, our generated images look much more natural. Besides, the more sophisticated sampling process in LMD makes the image generation slower. For instance, we observed that the sampling time of LMD is around 3$\times$ slower than our method.

\subsection{Blob Control over Using LLMs}

\begin{figure}[t]
    \centering
    \includegraphics[width=\textwidth]{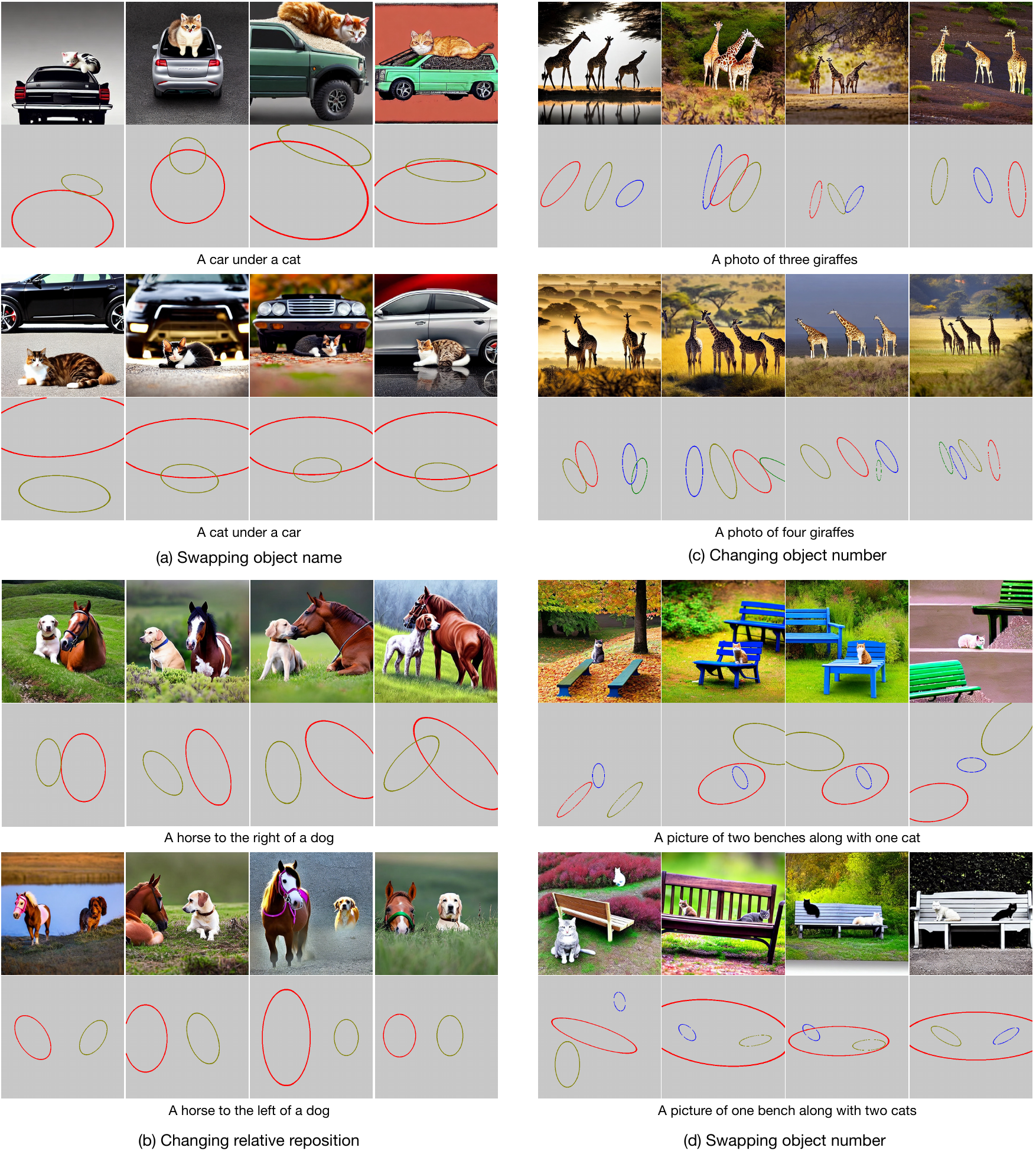}
    \vskip -0.1in
    \caption{Qualitative results of blob control over using LLMs, where we consider four cases of how LLMs understand compositional prompts for correct visual planning: (a) swapping object name (``cat" $\leftrightarrow$ ``car"), (b) changing relative reposition (``left" $\leftrightarrow$ ``right"), (c) changing object number (``three" $\leftrightarrow$ ``four"), and (d) swapping object number (``one bench \& two cats" $\leftrightarrow$ ``two benches \& one cat"). In each example, we show diverse blobs generated by LLMs (bottom) and the corresponding images generated by BlobGEN (top) from the same text prompt. Better zoom in for better visualization. }
    \label{fig:blobgen_llm_control}
\end{figure}

We add more qualitative results of blob control over using LLMs in Figure~\ref{fig:blobgen_llm_control}. To further demonstrate the blob control from LLMs, we consider four cases of how LLMs understand compositional prompts for correct visual planning: (a) swapping object name (``cat" $\leftrightarrow$ ``car"), (b) changing relative reposition (``left" $\leftrightarrow$ ``right"), (c) changing object number (``three" $\leftrightarrow$ ``four"), and (d) swapping object number (``one bench \& two cats" $\leftrightarrow$ ``two benches \& one cat"). We can see that LLMs have the ability of capturing the subtle differences when the prompts are modified in an ``adversarial" manner. Besides, we show that LLMs can generate diverse and feasible visual layouts from the same text prompt, which BlobGEN can use for synthesizing correct images of high quality and diversity.

\subsection{Failure Cases}

\begin{figure}[t]
    \centering
    \includegraphics[width=0.9\textwidth]{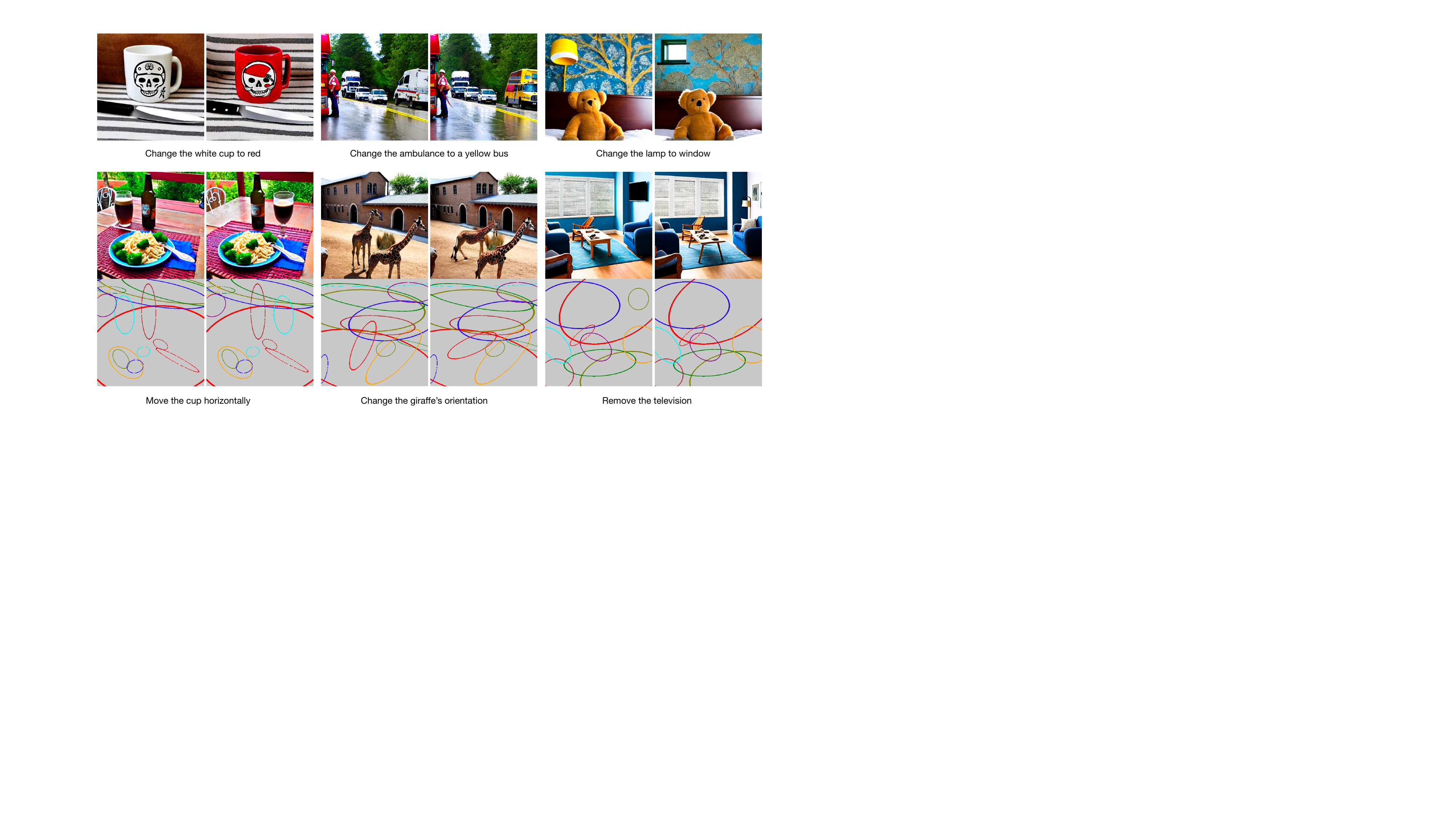}
    \vskip -0.1in
    \caption{Some failure cases for image editing. Similarly, each example contains two generated images: (Left) original setting and (Right) after editing. The top row shows the local editing results where we only change the blob description and since the blob parameters stay the same after editing, we do not show blob visualizations. The bottom two rows show the object reposition results where we only change the blob parameter. All images are in resolution of 512$\times$512. }
    \label{fig:coco_edit_fail_supp}
\end{figure}

\begin{figure}[t]
    \centering
    \includegraphics[width=0.9\textwidth]{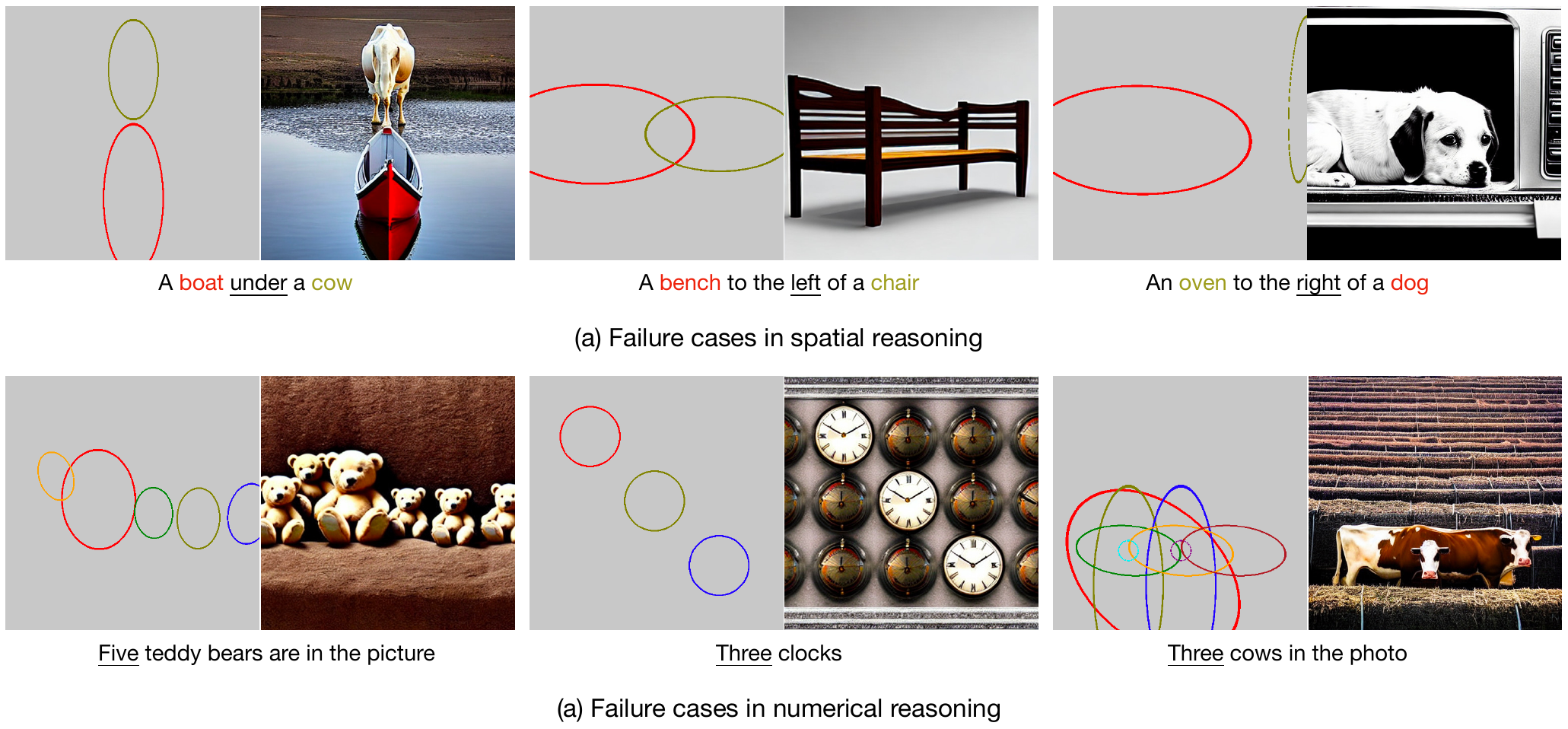}
    \vskip -0.1in
    \caption{Some failure cases for spatial and numerical reasoning. Given a caption, we prompt GPT4 to generate blob parameters (Left) and LLAMA-13B to generate blob descriptions (not shown in the figure), which are passed to our blob-grounded text-to-image generative model to synthesize an image (Right). All images are in resolution of 512$\times$512. }
    \label{fig:coco_reason_fail_supp}
\end{figure}

We show some failure cases when our method perform image editing and compositional reasoning tasks in Figures~\ref{fig:coco_edit_fail_supp} and \ref{fig:coco_reason_fail_supp}, respectively. For image editing, we see a few failure patterns: 1) the background also changes largely when the object to be edited is covered by the ``background'' blob, 2) the object itself changes when we only change its color or move its position, and 3) the edited object does not quite follow the instruction. We believe a combination with other image editing techniques will greatly reduce the failure rate. For compositional reasoning, we find that, on one hand, our method might not be perfectly robust to the LLM-generated blobs, and thus may have the ``copy-and-paste'' effect or completely fail when conditioning on some challenging or even wrong blobs. On the other hand, blob guidance does not prevent our method from generating more similar objects in other empty places, so our method may generate more objects than as instructed. 

\end{document}